\documentclass[journal]{IEEEtran}
\usepackage{cite}

\usepackage{threeparttable}

\ifCLASSINFOpdf
   \usepackage[pdftex]{graphicx}
   \DeclareGraphicsExtensions{.pdf,.jpeg,.png}
\else
   \usepackage[dvips]{graphicx}
   \DeclareGraphicsExtensions{.eps}
\fi
\usepackage{amsmath}
\usepackage{amsfonts}
\usepackage{amssymb}
\usepackage{multirow}
\usepackage{mathrsfs}
\usepackage{mathtools}
\usepackage{relsize}
\usepackage{bm}
\usepackage{times}
\usepackage{adjustbox}
\usepackage{mathptmx} 
\usepackage{textcomp}
\usepackage{array}
\usepackage{amsfonts}
\usepackage{amssymb}
\usepackage{multirow}
\usepackage{mathrsfs}
\usepackage{mathtools}
\usepackage{relsize}
\usepackage{bm}
\makeatletter
\newcommand{\removelatexerror}{\let\@latex@error\@gobble}
\makeatother

\usepackage{makecell}
\usepackage{algorithm}
\usepackage{algpseudocode}

\usepackage{array}

\ifCLASSOPTIONcompsoc
  \usepackage[caption=false,font=normalsize,labelfont=sf,textfont=sf]{subfig}
\else
  \usepackage[caption=false,font=footnotesize]{subfig}
\fi

\usepackage{stfloats}
\ifCLASSOPTIONcaptionsoff
 \usepackage[nomarkers]{endfloat}
 \let\MYoriglatexcaption\caption
 \renewcommand{\caption}[2][\relax]{\MYoriglatexcaption[#2]{#2}}
\fi
\let\MYorigsubfloat\subfloat
\renewcommand{\subfloat}[2][\relax]{\MYorigsubfloat[]{#2}}
\usepackage{url}

\begin{document}
%
\title{DeepCorrect: Correcting DNN Models against Image Distortions}
%
%
%

\author{Tejas~Borkar,~\IEEEmembership{Student~Member,~IEEE,}
        and~Lina~Karam,~\IEEEmembership{Fellow,~IEEE}%
\thanks{T. Borkar and L. Karam are with the Department
of Electrical, Computer and Energy Engineering, Arizona State University, Tempe,
AZ, 85281 USA e-mail: \{tsborkar, karam\}@asu.edu.}
}

\maketitle

\begin{abstract}
In recent years, the widespread use of deep neural networks (DNNs) has facilitated great improvements in performance for computer vision tasks like image classification and object recognition. In most realistic computer vision applications, an input image undergoes some form of image distortion such as blur and additive noise during image acquisition or transmission. Deep networks trained on pristine images perform poorly when tested on such distortions. In this paper, we evaluate the effect of image distortions like Gaussian blur and additive noise on the activations of pre-trained convolutional filters. We propose a metric to identify the most noise susceptible convolutional filters and rank them in order of the highest gain in classification accuracy upon correction. In our proposed approach called \emph{DeepCorrect}, we apply small stacks of convolutional layers with \emph{residual connections}, at the output of these ranked filters and train them to correct the worst distortion affected filter activations, whilst leaving the rest of the pre-trained filter outputs in the network unchanged. Performance results show that applying \emph{DeepCorrect} models for common vision tasks like image classification (ImageNet), object recognition (Caltech-101, Caltech-256) and scene classification (SUN-397), significantly improves the robustness of DNNs against distorted images and outperforms other alternative approaches.
\end{abstract}


%
\IEEEpeerreviewmaketitle

\section{Introduction}
%
%
%
%
\IEEEPARstart{T}{oday}, state-of-the-art algorithms for computer vision tasks like image classification, object recognition and semantic segmentation employ some form of deep neural networks (DNNs). The ease of design for such networks, afforded by numerous open source deep learning libraries \cite{caffe},\cite{keras}, has established DNNs as the go-to solution for many computer vision applications. Even challenging computer vision tasks like image classification \cite{vggnet},\cite{googlenet},\cite{he2016deep},\cite{AlexNet} and object recognition \cite{rcnn},\cite{fastrcnn},\cite{yolo}, which were previously considered to be extremely difficult, have seen great improvements in their state-of-the-art results due to the use of DNNs. An important factor contributing to the success of such deep architectures in computer vision tasks is the availability of large scale annotated datasets \cite{imagenet},\cite{coco}.

\begin{figure}[!t]
\centering
\subfloat[]{ \includegraphics[width=0.48\textwidth]{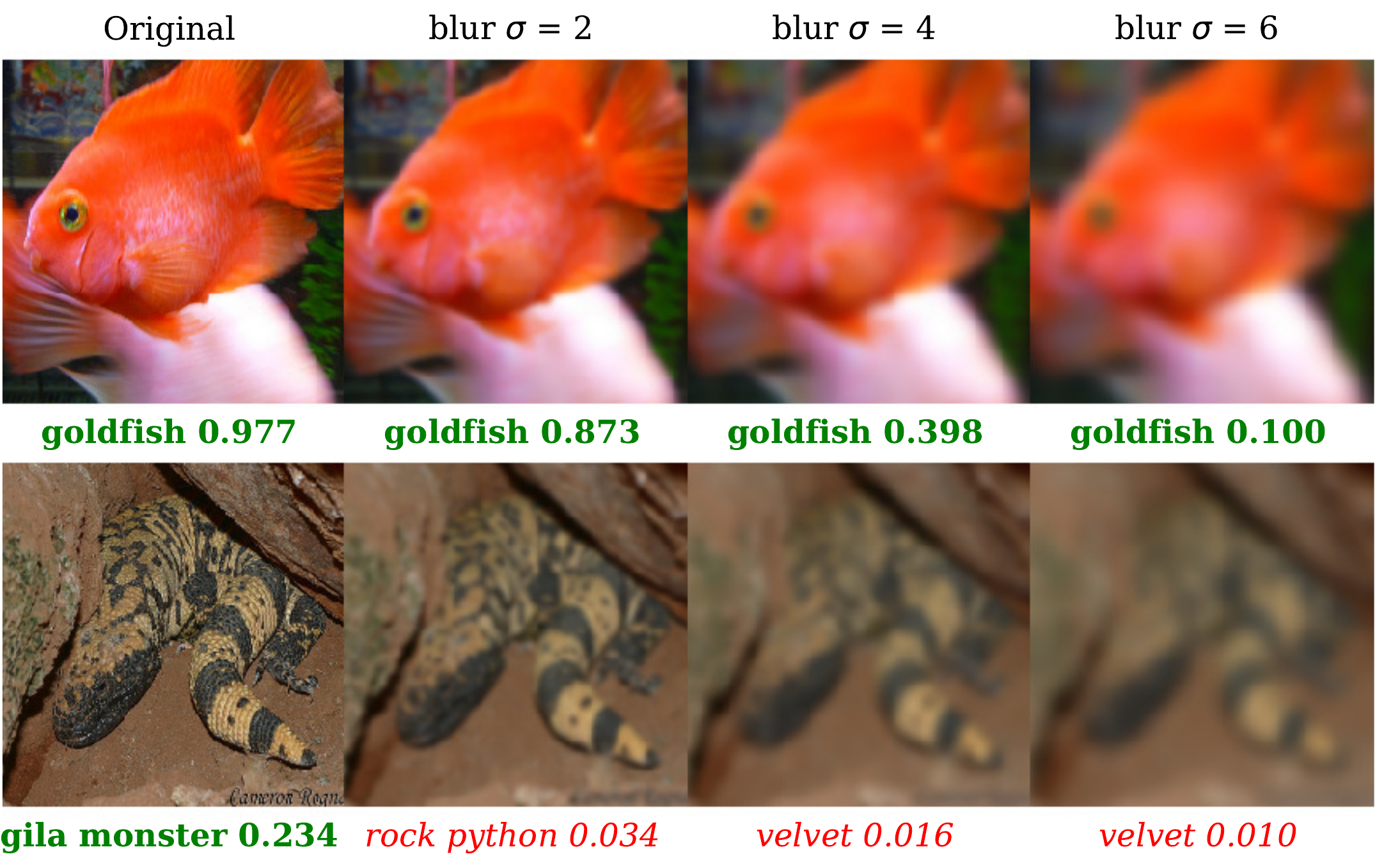}}
\label{fig:blur_intro}
\hfill
\vspace{-1pt}
\subfloat[]{\includegraphics[width=0.48\textwidth]{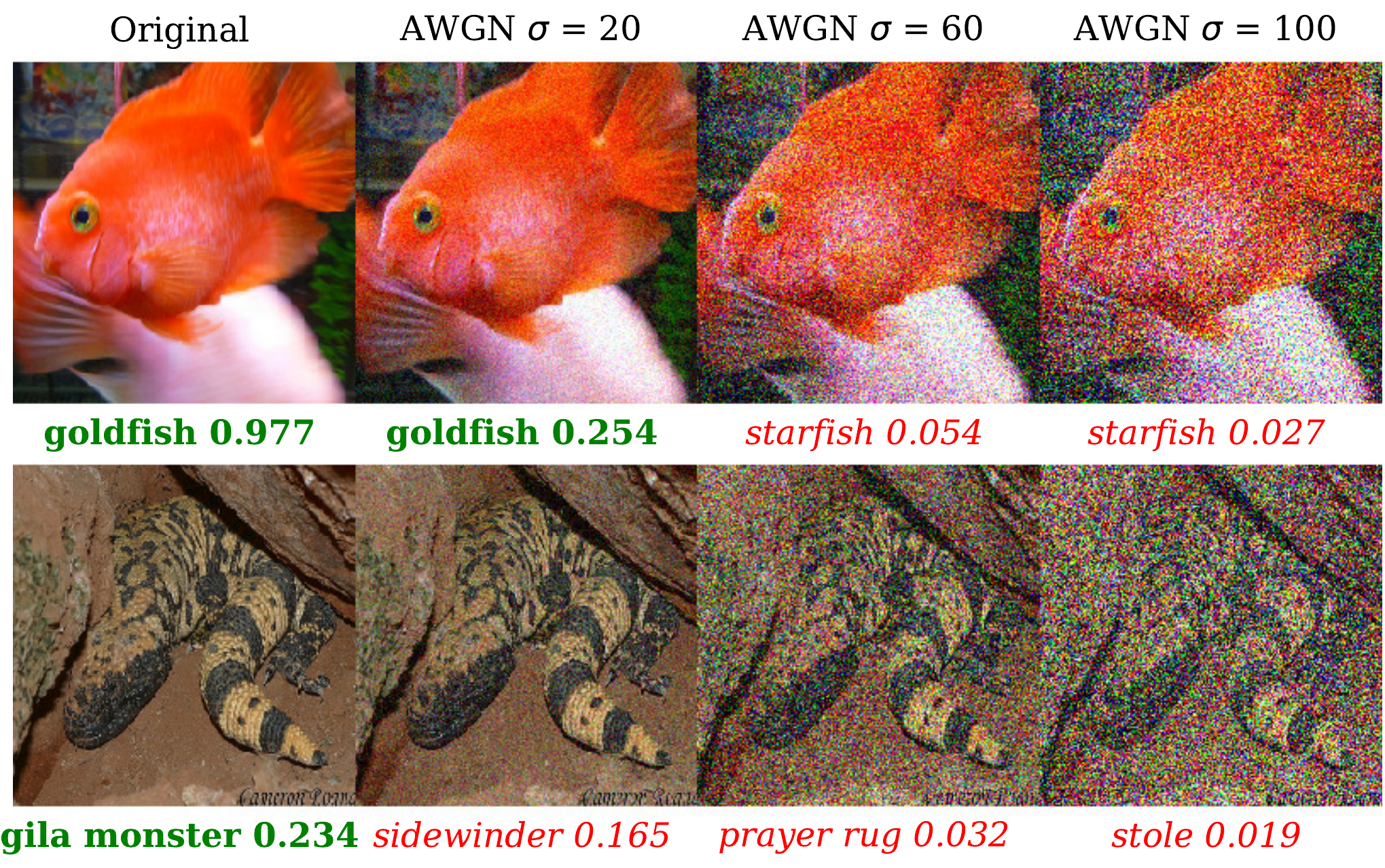}}
\label{fig:awgn_intro} 
\caption[]{Effect of image quality on DNN predictions, with predicted label and confidence generated by a pre-trained AlexNet \cite{AlexNet} model. Distortion severity increases from left to right, with the left-most image in a row having no distortion (original). Bold green text indicates correct classification, while italic red text denotes misclassification\protect\footnotemark[1].  (a) Examples from the ILSVRC2012 \cite{imagenet} validation set distorted by Gaussian blur. (b) Examples from the ILSVRC2012 validation set distorted by Additive White Gaussian Noise (AWGN) }
\vspace*{-3ex}
\label{fig:intro_fig}
\end{figure}

The visual quality of input images is an aspect very often overlooked while designing DNN based computer vision systems. In most realistic computer vision applications, an input image undergoes some form of image distortion including blur and additive noise during image acquisition, transmission or storage. However, most popular large scale datasets do not have images with such artifacts. Dodge and Karam \cite{sam} showed that even though such image distortions do not represent adversarial samples for a DNN, they do cause a considerable degradation in classification performance. \figurename~\ref{fig:intro_fig} shows the effect of image quality on the prediction performance of a DNN trained on high quality images devoid of distortions.  

Testing distorted images with a pre-trained DNN model for AlexNet \cite{AlexNet}, we observe that adding even a small amount of distortion to the original image results in a misclassification, even though the added distortion does not hinder the human ability to classify the same images (\figurename~\ref{fig:intro_fig}). In the cases where the predicted label for a distorted image is correct, the prediction confidence drops significantly as the distortion severity increases. This indicates that features learnt from a dataset of high quality images are not invariant to image distortion or noise and cannot be directly used for applications where the quality of images is different than that of the training images. Some issues to consider while deploying DNNs in noise/distortion affected environments include the following. For a network trained on undistorted images, are all convolutional filters in the network equally susceptible to noise or blur in the input image? Are networks able to learn some filters that are invariant to input distortions, even when such distortions are absent from the training set? Is it possible to identify and rank the convolutional filters that are most susceptible to image distortions and recover the lost performance, by only correcting the outputs of such ranked filters?



In our proposed approach called \emph{DeepCorrect}, we try to address these aforementioned questions through the following novel contributions: 
\begin{itemize}
    \item Evaluating the effect of common image distortions like Gaussian blur and Additive White Gaussian Noise (AWGN) on the outputs of pre-trained convolutional filters. We find that for every layer of convolutional filters in the DNN, certain filters are more susceptible to input distortions than others and that correcting the activations of these filters can help recover lost performance.
    \item  Measuring the susceptibility of convolutional filters to input distortions and ranking filters in the order of highest susceptibility to input distortion.
    \item Correcting the worst distortion-affected filter activations by appending \emph{correction units}, which are small blocks of stacked convolutional layers trained using a target-oriented loss, at the output of select filters, whilst leaving the rest of the pre-trained filter outputs in a DNN unchanged.
    \item Representing full-rank convolutions in our \emph{DeepCorrect} models with rank-constrained approximations consisting of a sequence of separable convolutions with rectangular kernels to mitigate the additional computational cost introduced by our \emph{correction units}. This results in pruned \emph{DeepCorrect} models that have almost the same computational cost as the respective pre-trained DNN, during inference. 
\end{itemize} 

Applying our \emph{DeepCorrect} models for common vision tasks like image classification \cite{imagenet}, object recognition \cite{caltech-101} \cite{caltech-256} and scene classification \cite{sun397} significantly improves the robustness of DNNs against distorted images and also outperforms other alternative approaches, while training significantly lesser parameters. To ensure reproducibility of presented results, the code for \emph{DeepCorrect} is made publicly available at \url{https://github.com/tsborkar/DeepCorrect}.

The remainder of the paper is organized as follows. Section \ref{sec:related_work} provides an overview of the related work in assessing and improving the robustness of DNNs to input image perturbations. Section \ref{sec:baseline} describes the distortions, network architectures and datasets we use for analyzing the distortion susceptibility of convolutional filters in a DNN.  A detailed description of our proposed approach is presented in Section \ref{sec:dc} followed, in Section \ref{sec:exp}, by extensive experimental validation with different DNN architectures and multiple datasets covering image classification, object recognition and scene classification. Concluding remarks are given in Section~\ref{sec:conc}.
\protect\footnotetext[1]{All figures in this paper are best viewed in color.}
\vspace*{-3ex}
\section{Related Work}
\label{sec:related_work}
DNN susceptibility to specific small magnitude perturbations which are imperceptible to humans but cause networks to make erroneous predictions with high confidence (adversarial samples) has been studied in \cite{szegedy2013intriguing} and \cite{adverserial_samples}. The concept of rubbish samples proposed in \cite{fooling_dnn} studies the vulnerability of DNNs to make arbitrary high confidence predictions for random noise images that are completely unrecognizable to humans, i.e., the images contain random noise and no actual object. However, both adversarial samples and rubbish samples are relatively less encountered in common computer vision applications as compared to other common distortions due to image acquisition, storage, transmission and reproduction.


Karam and Zhu \cite{qlfw} present QLFW, a face matching dataset consisting of images with five types of quality distortions. Basu \textit{et al.} \cite{nmnist} present the n-MNIST dataset, which adds Gaussian noise, motion blur and reduced contrast to the original images of the MNIST dataset. Dodge and Karam~\cite{sam} evaluate the impact of a variety of quality distortions such as Gaussian blur, AWGN and JPEG compression on various state-of-the-art DNNs and report a substantial drop in classification accuracy on the ImageNet (ILSVRC2012) dataset in the presence of blur and noise. A similar evaluation for the task of face recognition is presented in \cite{face_quality}.

Rodner \textit{et al.} \cite{rodner} assess the sensitivity of various DNNs to image distortions like translation, AWGN and salt \& pepper noise, for the task of fine grained categorization on the CUB-200-2011 \cite{cub200} and Oxford flowers \cite{oxford_flowers} datasets, by proposing a first-order Taylor series based gradient approximation that measures the expected change in final layer outputs for small perturbations to input image. Since a gradient approximation assumes small perturbations, Rodner \textit{et al.}'s sensitivity measure does not work well for higher levels of distortion as shown by \cite{rodner} and does not assess susceptibility at a filter level within a DNN. Furthermore, Rodner \textit{et al.} do not present a solution for making the network more robust to input distortions; instead, they simply fine-tune the whole network with the distorted images added as part of data augmentation during training. Retraining large networks such as VGG16 \cite{vggnet} or ResNet-50 \cite{he2016deep} on large-scale datasets is computationally expensive. Unlike Rodner \textit{et al.}'s work, our proposed ranking measure assesses sensitivity of individual convolutional filters in a DNN, is not limited to differentiable distortion processes, and holds good for both small and large perturbations.

Zheng \textit{et al.} \cite{stability} improve the general robustness of DNNs to unseen small perturbations in the input image through the introduction of distortion agnostic \emph{stability training}, which minimizes the KL-divergence between DNN output predictions for a clean image and a noise perturbed version of the same image. The perturbed images are generated by adding uncorrelated Gaussian noise to the original image. \emph{Stability training} provides improved DNN robustness against JPEG compression, image scaling and cropping. However, the top-1 accuracy of \emph{stability trained} models is lower than the original model, when tested on most distortions including motion blur, defocus blur and additive noise among others~\cite{feat_quantize}. Sun \textit{et al.} also propose a distortion agnostic approach to improve DNN robustness by introducing 3 feature quantization operations, i.e., a floor function with adaptive resolution, a power function with learnable exponents and a power function with data dependent exponents, that act on the convolutional filter activations before the ReLU non-linearity is applied. However, similar to \emph{stability training}, the performance of \emph{feature quantized} models is lower than the original model, when tested on distortions like defocus blur and additive noise. Additionally, no single feature quantization function consistently outperforms the other two for all types of distortion. Although both distortion agnostic methods \cite{stability}, \cite{feat_quantize} improve DNN robustness, their top-1 accuracy is much lower than the accuracy of the original DNN on distortion free images, making it difficult to deploy these models. 

A non-blind approach to improve the resilience of networks trained on high quality images would be to retrain the network parameters (fine-tune) on images with observed distortion types. Vasiljevic \textit{et al.} \cite{fine_tune} study the effect of various types of blur on the performance of DNNs and show that DNN performance for the task of classification and segmentation drops in the presence of blur. Vasiljevic \textit{et al.} \cite{fine_tune} and Zhou \textit{et al.} \cite{zhou2017classification} show that fine-tuning a DNN on a dataset comprised of both distorted and undistorted images helps to recover part of the lost performance when the degree of distortion is low. 

Diamond \textit{et al.} \cite{dirty_pixel} propose a joint denoising, deblurring and classification pipeline. This involves an image preprocessing stage that denoises and deblurs the image in a manner that preserves image features optimal for classification rather than aesthetic appearance. The classification stage has to be fine-tuned using distorted and clean images, while the denoising and deblurring stages assume a priori knowledge of camera parameters and the blur kernel, which may not be available at the time of testing.   

\begin{figure}[]
\centering
\subfloat[(a)]{\includegraphics[width=0.45\linewidth, height=5.5in]{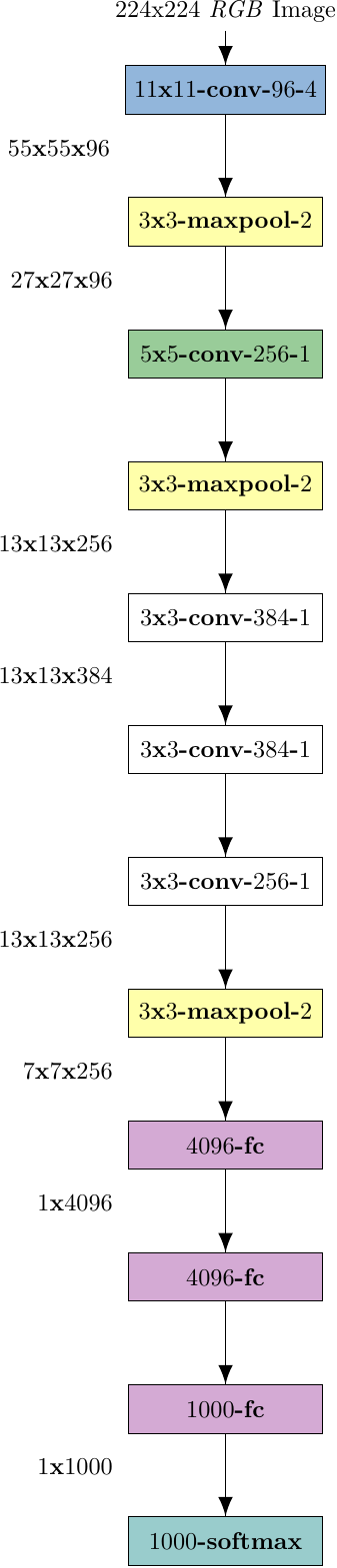}
\label{fig:AlexNet}
}
\hspace{10pt}
\subfloat{\includegraphics[width=0.45\linewidth, height=5.5in]{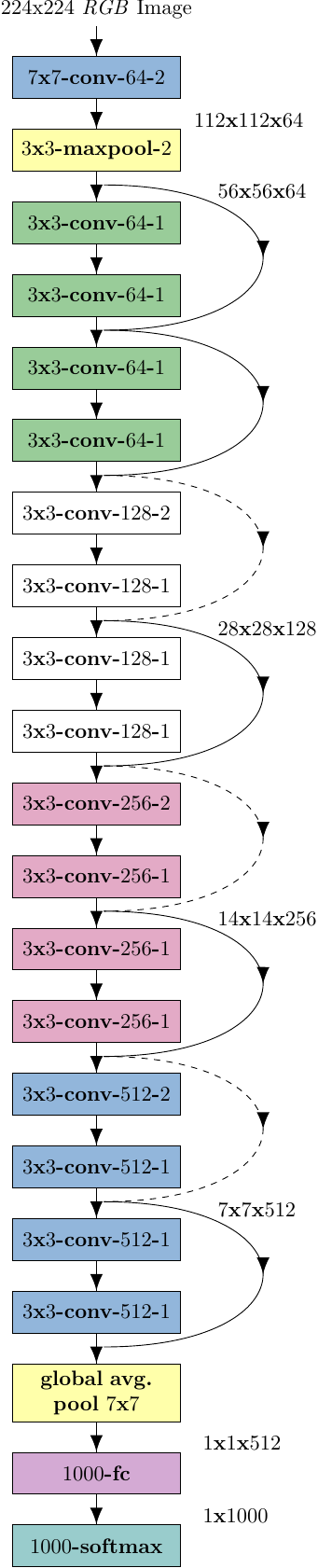}
\label{fig:res18}
}
\caption{ Network architectures for our baseline models. Convolutional layers are parameterized by $k$x$k$-conv-$d$-$s$, where $k$x$k$ is the spatial extent of the filter, $d$ is the number of output filters in a layer and $s$ represents the filter stride. Maxpooling layers are parameterized as $k$x$k$-maxpool-$s$, where $s$ is the spatial stride. (a) AlexNet DNN \cite{AlexNet}\protect\footnotemark[3] (b) ResNet18 \cite{he2016deep}\protect\footnotemark[4].}
\vspace*{-2ex}
\label{fig:base_net}
\end{figure}

\vspace*{-4ex}
\section{Experimental Setup}
\label{sec:baseline}
\vspace*{-2ex}
Here, we describe the various image distortions and network architectures used to evaluate the susceptibility of individual convolutional filters to input distortions. We use the ILSVRC-2012 validation set \cite{imagenet} for our experiments.
\vspace*{-6ex}
\subsection{Distortions} 
\vspace*{-2ex}
We first focus on evaluating two important and conflicting types of image distortions: Gaussian blur and AWGN over 6 levels of distortion severity. Gaussian blur, often encountered during image acquisition and compression \cite{tid2008}, represents a distortion that eliminates high frequency discriminative object features like edges and contours, whereas AWGN is commonly used to model additive noise encountered during image acquisition and transmission. We use a noise standard deviation $\sigma_n \in $ \{10, 20, 40, 60, 80, 100\} for AWGN and blur standard deviation $\sigma_b \in $ \{1, 2, 3, 4, 5, 6\} for Gaussian blur. The size of the blur kernel is set to 4 times the value of $\sigma_b$.
\protect\footnotetext[2]{We use the code and weights for the AlexNet DNN available online at \url{https://github.com/heuritech/convnets-keras}}
\protect\footnotetext[3]{Every convolutional layer is followed by a ReLU non-linearity for the AlexNet DNN. In addition to the ReLU non-linearity, the first and second convolutional layers of the AlexNet DNN are also followed by a local response normalization operation \cite{AlexNet}.}
\protect\footnotetext[4]{Every convolutional layer is followed by a batch normalization operation and ReLU non-linearity for the ResNet18 model. Skip connections and residual feature maps are combined through an element-wise addition. Dashed-line skip connections perform an identity mapping using a stride of 2 to reduce feature map size and pad zero entries along the channel axis to increase dimensionality \cite{he2016deep}.
\vspace*{-3ex}}
\vspace*{-8pt}
\subsection{Network Architectures} 
 We use two different network architectures as shown in \figurename~\ref{fig:base_net}, specifically: a shallow 8-layered DNN (AlexNet)\protect\footnotemark[2] and a deeper 18-layered network with residual connections (ResNet18) \cite{he2016deep}. We use the term "pre-trained" or "baseline" network to refer to any network that is trained on undistorted images.

\section{DeepCorrect}
\label{sec:dc}
Although pre-trained networks perform poorly on test images with significantly different image statistics than those used to train these networks (Table \ref{baseline_res_blur}), it is not obvious if only some convolutional filters in a network layer are responsible for most of the observed performance gap or if all convolutional filters in a layer contribute more or less equally to the performance degradation. If only a subset of the filters in a layer are responsible for most of the lost performance, we can focus on restoring only the most severely affected activations and avoid modifying all the remaining filter activations in a DNN. 
\begin{table}[!t]
\scriptsize
\renewcommand{\arraystretch}{1.3}
\setlength{\tabcolsep}{1.2em}
\caption{Top-1 accuracy of pre-trained networks for distortion affected images as well as undistorted images (original). For Gaussian blur and AWGN, accuracy is reported by averaging over all levels of distortion.}
\vspace*{-5pt}
\label{baseline_res_blur}            
\centering
\begin{tabular}{|c|c|c|c|}
\hline
Models & Original & Gaussian blur & AWGN \\ \hline
AlexNet &  0.5694 & 0.2305 & 0.2375 \\
ResNet18 & 0.6912  & 0.3841 &0.3255 \\
\hline
\end{tabular}
\vspace*{-3ex}
\end{table}
\normalsize
\vspace*{-4pt}
\subsection{Ranking Filters through Correction Priority}
\label{subsec:ranking}


We define the output of a single convolutional filter $\phi_{i,j}$ to the input ${\bf{x}}_i$  by $\phi_{i,j}({\bf{x}}_i)$,  where $i$ and $j$ correspond to layer number and filter number, respectively. If $g_{i}(\cdot)$ is a transformation that models the distortion acting on filter input ${\bf{x}}_i$, then the output of a convolutional filter $\phi_{i,j}$ to the distortion affected input is given by $\widetilde{\phi_{i,j}}({\bf{x}}_i) = \phi_{i,j}(g_i({\bf{x}}_i))$. It should be noted that $\widetilde{\phi_{i,j}} ({\bf{x}}_i)$ represents the filter activations generated by distorted inputs and $\phi_{i,j}({\bf{x}}_i)$ represents the filter activations for undistorted inputs. Assuming we have access to $\phi_{i,j}({\bf{x}}_i)$ for a given set of input images, replacing $\widetilde{\phi_{i,j}}({\bf{x}}_i)$ with $\phi_{i,j}({\bf{x}}_i)$ in a deep network is akin to perfectly correcting the activations of the convolutional filter $\phi_{i,j}$ against input image distortions. Computing the output predictions by swapping a distortion affected filter output with its corresponding clean output for each of the ranked filters would improve classification performance. 
The extent of improvement in performance is indicative of the susceptibility of a particular convolutional filter to input distortion and its contribution to the associated performance degradation. 
\begin{figure}[]
\begin{algorithm}[H]
\caption{Computing Correction Priority} \label{alg:ranking}
\begin{algorithmic}[1]
\small
\renewcommand{\algorithmicrequire}{\textbf{Input:}}
 \renewcommand{\algorithmicensure}{\textbf{Output:}}
\Require $({\bf{x}}_{1,i}, g_i({\bf{x}_{1,i}}), y_{1}),\ldots, ({\bf{x}}_{M,i}, g_i({\bf{x}_{M,i}}), y_{M})$ are given triplets with $ 1\le i\le L$, where $i$ represents the layer number, ${\bf{x}_{m,i}}$ is the $m^{th}$ undistorted input for layer $i$ and $g_i({\bf{x}_{m,i}})$ is the corresponding distorted version, $M$ is the total number of images in the validation set and $y_m$ is the ground-truth label for the $m^{th}$ input image.
\Ensure Correction priority $\tau$
\State {$p_b := 0$}
\For  {$m = 1$ to $M$}
\State {Predict class label $ypred_{m}$ for distorted image ${g_1(\bf{x}}_{m,1})$}
\State {Compute $p_b = p_b + \frac{1}{M}h(y_m,ypred_m)$, \\ where $h(y_m,ypred_m) =$ 1, if $y_m = ypred_m$ and 0 otherwise.}
\EndFor
\For{$i = 1$ to L}
 \State $N_{i} \gets$ number of filters in layer $i$
 \State $\phi_{i,j} \gets$ $j^{th}$ convolutional filter in the $i^{th}$ layer
\For{$j = 1$ to $N_{i}$}
 \State $p_{swp}(j)$ = 0
   \For{$m = 1$ to $M$}
    \State $\phi_{i,j}(g_i({\bf{x}}_{m,i})) \gets  \phi_{i,j}({\bf{x}}_{m,i})$
    \State Predict class label $ypred_{m}$ 
     \State {$p_{swp}(j) = p_{swp}(j) + \frac{1}{M}h(y_m,ypred_m)$,\\ where $h(y_m,ypred_m) =$ 1, if $y_m = ypred_m$ and 0 otherwise.}
    \EndFor
\EndFor
\State $\tau(i,j) \gets p_{swp}(j) - p_b$
\EndFor
\State \textbf{return} $\tau$
\end{algorithmic}
\end{algorithm}
\vspace*{-6ex}
\end{figure}
\normalsize
\begin{figure*}[]
	\centering
	\includegraphics[width=\linewidth]{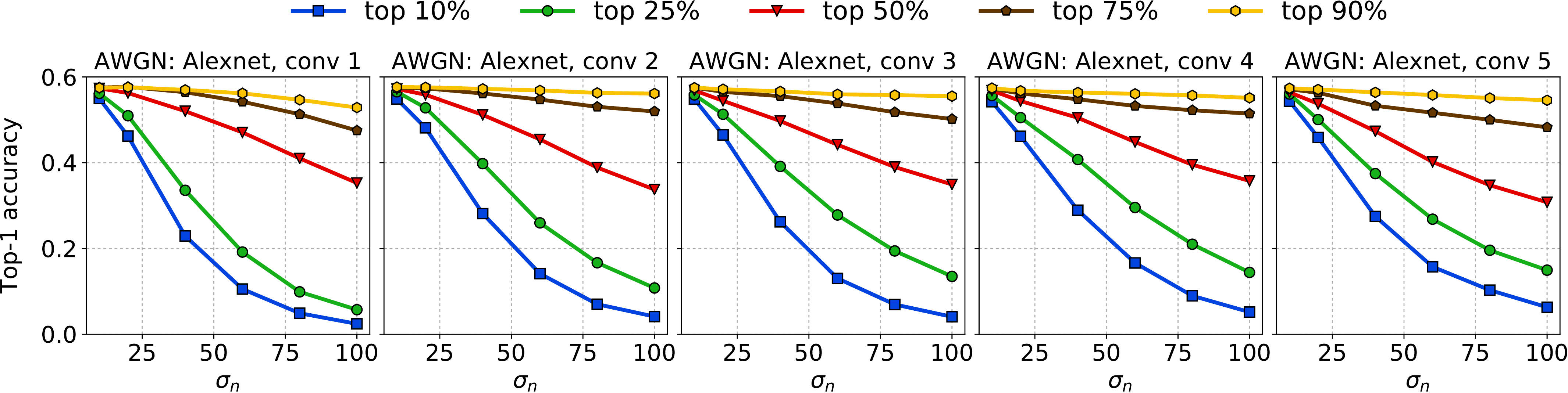}	
		\hfill
	\includegraphics[width=\linewidth]{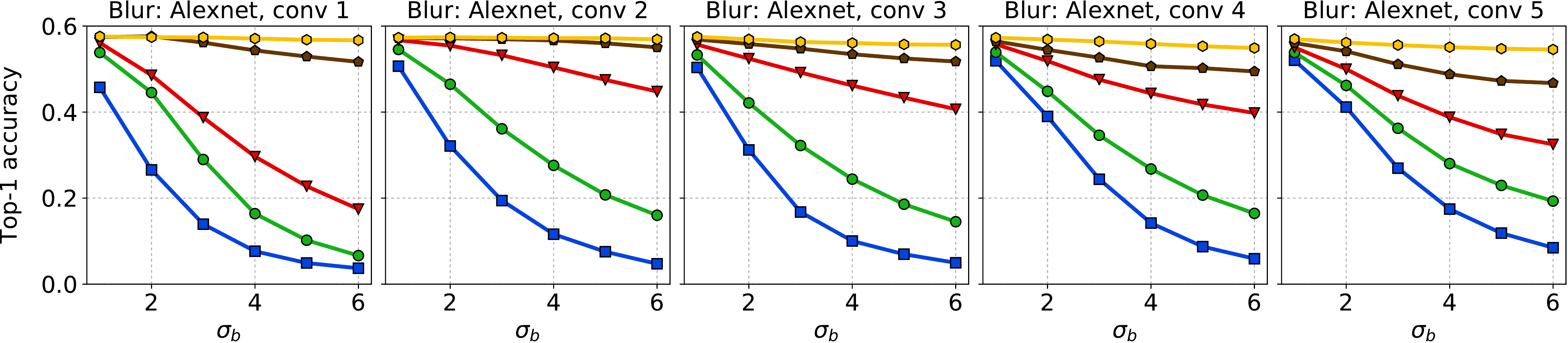}
	\caption{Effect of varying the percentage of corrected filter activations $\beta_i \in$ \{10\%, 25\%, 50\%, 75\%, 90\%\}, in the $i^{th}$ convolutional layer (conv $i$) of pre-trained AlexNet, for AWGN and Gaussian blur affected images, respectively.}
	\label{fig:theoretical_res_corr}
	\vspace*{-3ex}
\end{figure*}

We now define the correction priority of a convolutional filter $\phi_{i,j}$  as the improvement in DNN performance on a validation set, generated by replacing $\widetilde{\phi_{i,j}} ({\bf{x}}_i)$ with $\phi_{i,j}({\bf{x}}_i)$  for a pre-trained network. Let the baseline performance (computed over distorted images) for a network be $p_{b}$, which can be obtained by computing the average top-1 accuracy of the network over a set of images or another task-specific performance measure. Let $p_{swp}(i,j)$ denote the new improved performance of the network after swapping $\widetilde{\phi_{i,j}} ({\bf{x}}_i)$ with $\phi_{i,j}({\bf{x}}_i)$. As our implementation focuses on classification tasks, the average top-1 accuracy over a set of distorted images is used to measure $p_{b}$ and $p_{swp}(i,j)$. The correction priority for filter $\phi_{i,j}$ is then given by:
\vspace*{-4pt}
\begin{equation}
\tau(i,j) = p_{swp}(i, j)-p_b
\label{equ:norm_corr_priority}
\end{equation}
\normalsize
A higher $\tau(i,j)$ indicates higher susceptibility of the convolutional filter $\phi_{i,j}$ to input distortion. 
Using the proposed ranking measure in Equation (\ref{equ:norm_corr_priority}) and 5000 images (i.e., 5 images per class) randomly sampled from the ILSVRC-2012 training set, we compute correction priorities for every convolutional filter in the network and rank the filters in descending order of correction priority. The detailed overview and pseudo-code for computing correction priorities is summarized in Algorithm \ref{alg:ranking}. 
 
We evaluate in \figurename~\ref{fig:theoretical_res_corr}, the effect of correcting different percentages, $\beta_i$, of the ranked filter activations in the $i^{th}$ DNN layer of AlexNet for distortion affected images. For the AlexNet model, it is possible to recover a significant amount of the lost performance, by correcting only 50\% of the filter activations in any one layer, which indicates that a select subset of convolutional filters in each layer are indeed more susceptible to distortions than the rest. Although we show graphs for the AlexNet model, form our experiments, we make similar observations for the ResNet18 model as well.

\begin{figure}[!t]
\begin{minipage}{\linewidth}
\centering
\subfloat[(a)]{\includegraphics[width=0.9\linewidth]{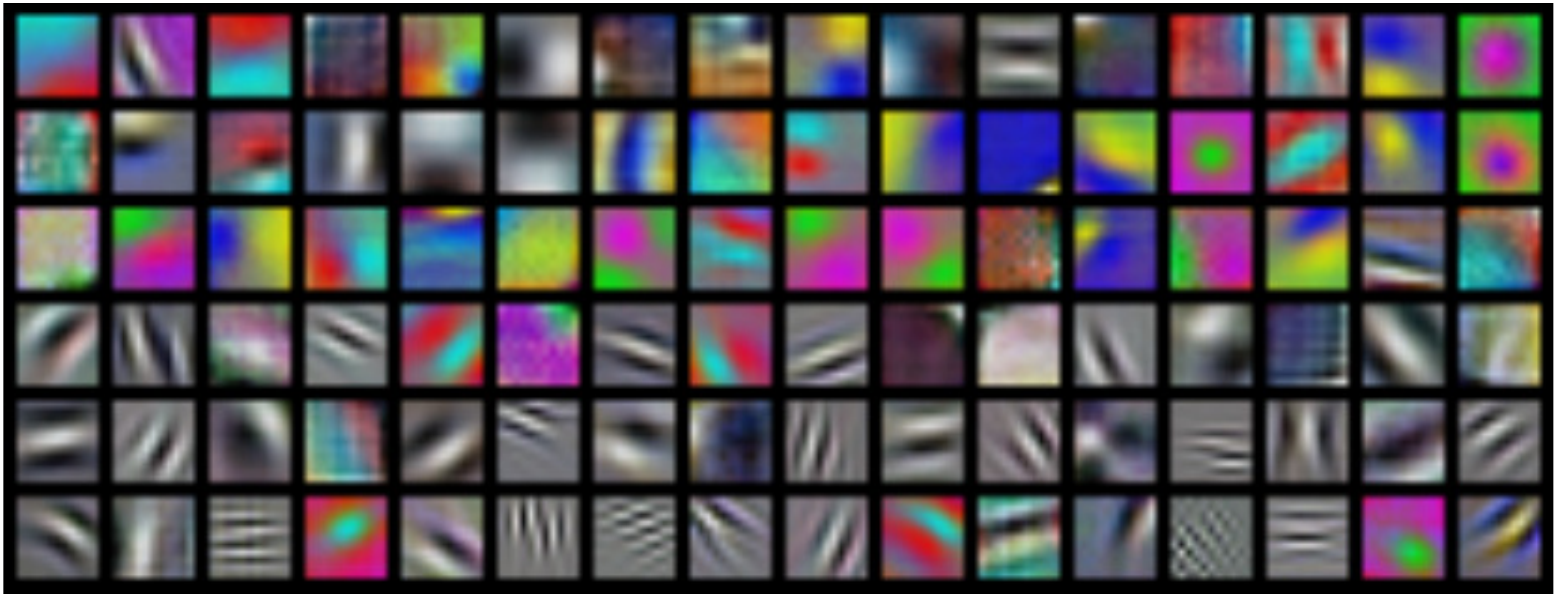}
\label{fig:ly1_filt}}
\vspace{-1pt}
\centering
\subfloat[(b)]{\includegraphics[width=0.9\linewidth]{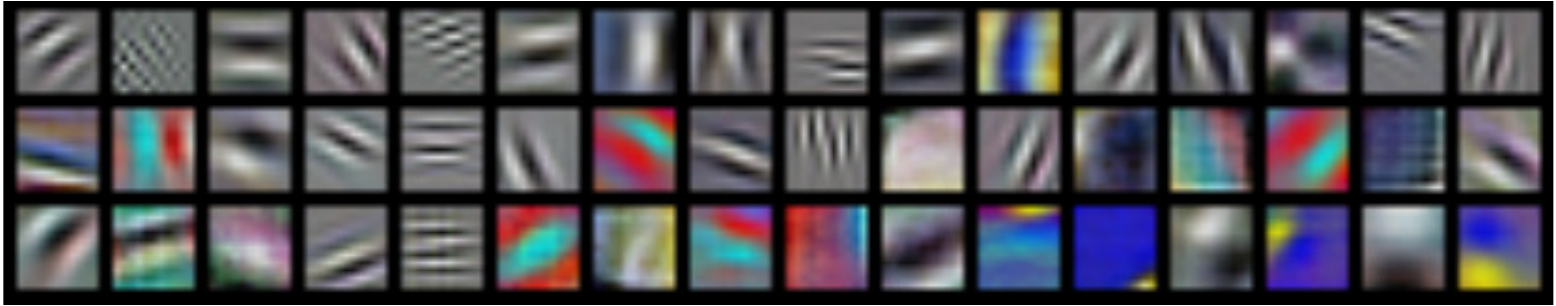}
\label{fig:top_50_blur}}
\vspace{-1pt}
\centering
\subfloat[(c)]{\includegraphics[width=0.9\linewidth]{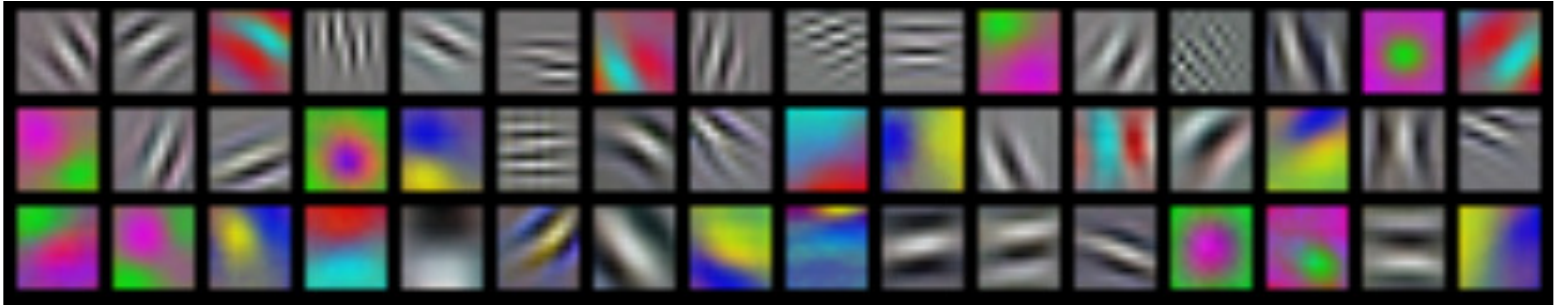}
\label{fig:top_50_awgn}}
\caption[]{(a) 96 convolutional filter kernels of size 11x11x3 in the first convolutional layer of pre-trained AlexNet. (b) Convolutional filter kernels most susceptible to Gaussian blur (top 50\%), as identified by our proposed ranking metric. (c) Convolutional filter kernels most susceptible to AWGN (top 50\%), as identified by our proposed ranking metric. In (b) and (c), the filters are sorted in descending order of susceptibility going row-wise from top left to bottom right.}
\label{fig:filt_viz}
\vspace*{-3ex}
\end{minipage}
\end{figure}

Convolutional filter visualizations from the first layer of the pre-trained AlexNet model (\figurename ~\ref{fig:ly1_filt}) reveal two types of filter kernels: 1) mostly color agnostic, frequency- and orientation-selective filters that capture edges and object contours and 2) color specific blob shaped filters that are sensitive to specific color combinations. Figs.~\ref{fig:top_50_blur} and \ref{fig:top_50_awgn} visualize the top 50\% filters in the first convolutional layer of AlexNet, that are most susceptible to Gaussian blur and AWGN, respectively, as identified by our proposed ranking metric. The identified filters most susceptible to Gaussian blur are mainly frequency- and orientation-selective filters, most of which are color agnostic, while filters most susceptible to AWGN are a mix of both color specific blobs and frequency- and orientation-selective filters. This is in line with our intuitive understanding that Gaussian blur majorly affects edges and object contours and not object color, while AWGN affects color as well as object contours.  

\subsection{Correcting Ranked Filter Outputs}
\label{subsec:corr_unit} 
\begin{figure}[!t]
\centering
\includegraphics[width=0.35\linewidth]{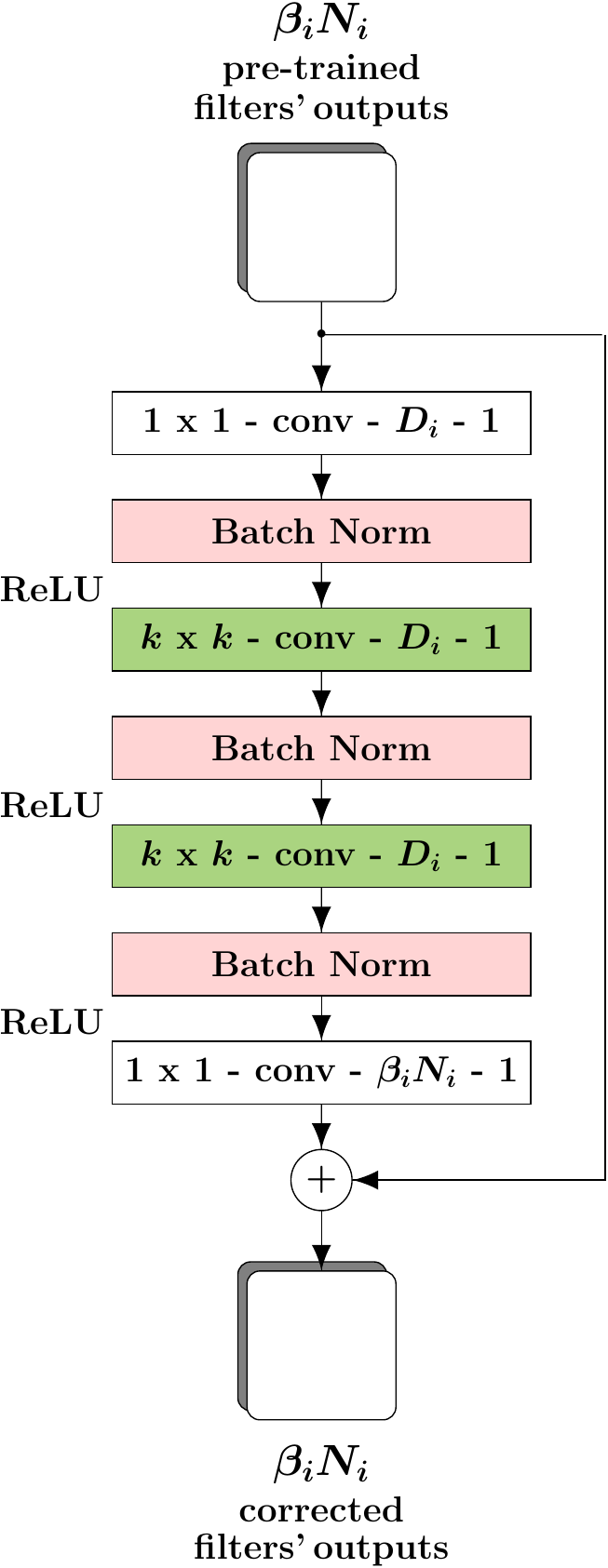}
\caption{\emph{Correction unit} based on a residual function~\cite{he2016deep}, acting on the outputs of $\beta_iN_i$ $(0 <\beta_i< 1)$ filters out of $N_i$ total filters in the $i^{th}$ convolutional layer of a pre-trained DNN. All convolutional layers in the \emph{residual block}, except the first and last layer, are parameterized by $k$x$k$-conv-$D_i$-$s$, where $k$x$k$ is spatial extent of the filter, $D_i$ (correction unit kernel depth) is the number of output filters in a layer, $s$ represents the filter stride and $i$ represents the layer number of the convolutional layer being corrected in the pre-trained DNN.}
\label{fig:corr_unit}
\vspace*{-2ex}
\end{figure}

Here, we propose a novel approach, which we refer to as \emph{DeepCorrect}, where we learn a task-driven corrective transform that acts as a distortion masker for convolutional filters that are most susceptible to input distortion, while leaving all the other pre-trained filter outputs in the layer unchanged. Let $R_i$ represent a set consisting of the $N_i$ ranked filter indices in the $i^{th}$ layer of the network, computed using the procedure in Section \ref{subsec:ranking}. Also let $R_{i,\beta_i}$ represent a subset of $R_i$ consisting of the top $\beta_iN_i$ ranked filter indices in network layer $i$, where $N_i$ is the total number of convolutional filters in layer $i$ and $\beta_i$ is the percentage of filters corrected in layer $i$, as defined in Section \ref{subsec:ranking}. If $\Phi_i$ represents the set of convolutional filters in the $i^{th}$ layer, the objective is to learn a transform $F_{corr_i}(:)$ such that:
\begin{equation}
F_{corr_i}(\Phi_{R_{i,\beta_i}}(g_i({\bf{x}}_i))) \approx \Phi_{R_{i,\beta_i}}({\bf{x}}_i)
\end{equation}
where ${{\bf{x}}_i}$ is the undistorted input to the $i^{th}$ layer of convolutional filters and $g_{i}(\cdot)$ is a transformation that models the distortion acting on ${{\bf{x}}_i}$. Since we do not assume any specific form for the image distortion process, we let the corrective transform $F_{corr_i}(:)$ take the form of a shallow \emph{residual block}, which is a small stack of convolutional layers (4 layers) with a single skip connection~\cite{he2016deep}, such as the one shown in \figurename~\ref{fig:corr_unit}. We refer to such a \emph{residual block} as a \emph{correction unit}. $F_{corr_i}(:)$ can now be estimated using a target-oriented loss such as the one used to train the original network, through backpropogation~\cite{backprop}, but with much less number of parameters. 
\begin{figure}[!t]
\centering
\includegraphics[height=5in, width=0.51\linewidth]{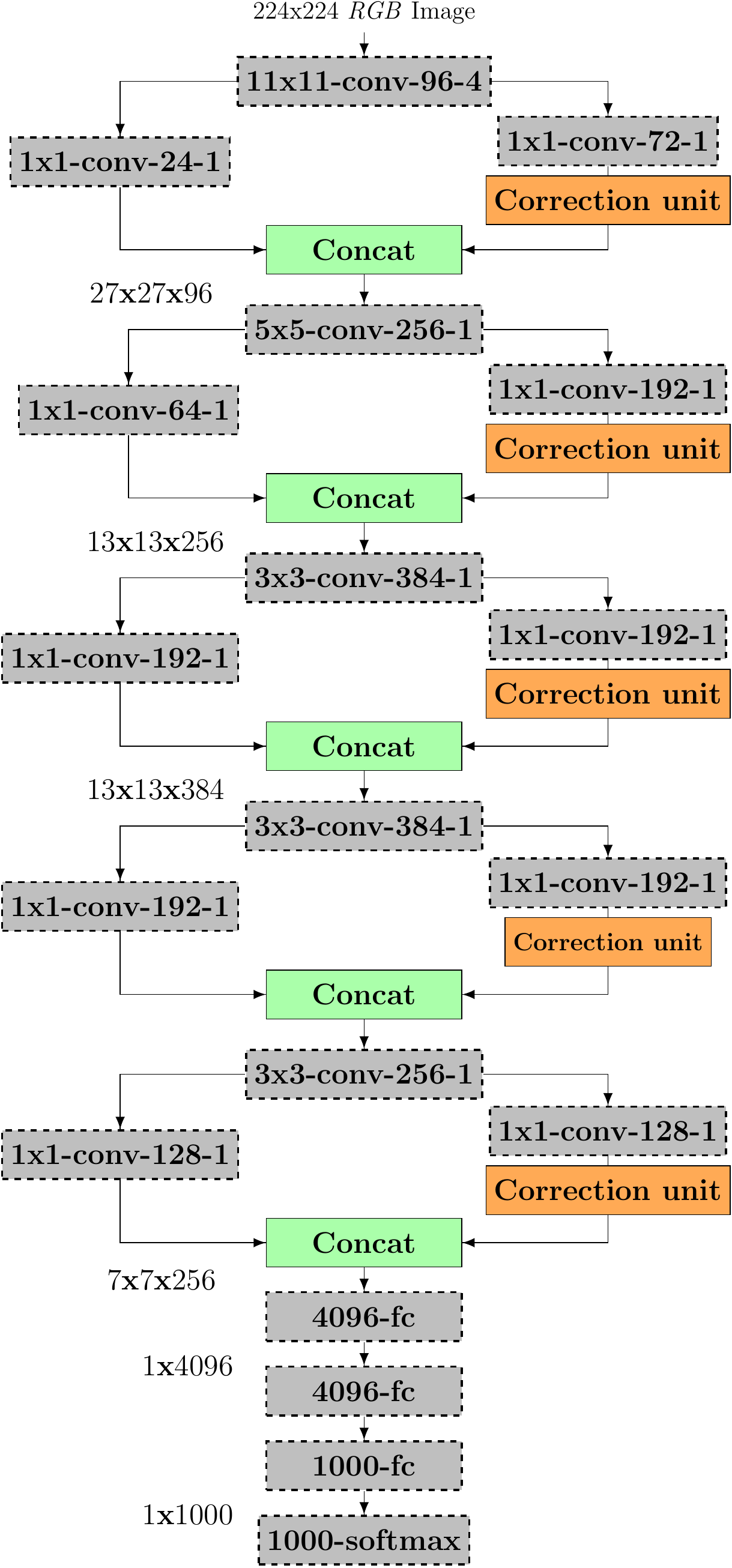}
\label{fig:AlexNet_corr}
\caption{\emph{DeepCorrect} model for AlexNet, with 75\% filter outputs corrected in the first two layers and 50\% filter outputs corrected in the next three layers\protect\footnotemark[5]. Convolution layers from the original architecture in \figurename~\ref{fig:base_net}, shown in gray with dashed outlines, are non-trainable layers and their weights are kept the same as those of the pre-trained model.}
\label{fig:deep_corr}
\vspace*{-2ex}
\end{figure}

Consider an $L$ layered DNN $\Phi$ that has been pre-trained for an image classification task using clean images. $\Phi$ can be interpreted as a function that maps network input ${\bf{x}}$ to an output vector $\Phi({\bf{x} }) \in \mathbb{R}^d$, such that:
\begin{equation}
\Phi = \Phi_{L}\circ\Phi_{L-1}\circ\ldots\Phi_{2}\circ\Phi_{1}
\end{equation}
where $\Phi_{i}$ is the mapping function (set of convolutional filters) representing the $i^{th}$ DNN layer and $d$ is the dimensionality of the network output. 

Without loss of generality, if we add a \emph{correction unit} that acts on the top $\beta_1N_1$ ranked filters in the first network layer, then the resultant network $\Phi_{corr}$ is given by: 
\begin{equation}
\Phi_{corr} = \Phi_{L}\circ\Phi_{L-1}\circ\ldots\Phi_{2}\circ\Phi_{1_{corr}}
\end{equation}
where $\Phi_{1_{corr}}$ represents the new mapping function for the first layer, in which the corrective transform $F_{corr_1}(:)$ acts on the activations of the filter subset $\Phi_{R_{1,\beta_1}}$ and all the remaining filter activations are left unchanged. If ${\bf{W}}_{1}$ represents the trainable parameters in $F_{corr_{1}}$, then  $F_{corr_{1}}$ can be estimated by minimizing :
\begin{equation}
\small
E({\bf{W}}_{1}) = \lambda\mathcal{R}({\bf{W}_{1}}) + \frac{1}{M}\mathlarger{\sum}\limits_{m=1}^{M}\mathcal{L}(y_{m},\Phi_{corr}({\bf{x}}_{m}))
\label{equ:target_loss}
\end{equation}
\normalsize
where $\lambda$ is a constant, $\mathcal{R}$ is a regularizer such as $l_1$ norm or $l_2$ norm, $\mathcal{L}$ is a standard cross-entropy classification loss, $y_m$ is the target output label for the $m^{th}$ input image ${\bf{x}}_{m}$, $M$ represents the total number of images in the training set and, since we train on a collection of both distorted and clean images, ${\bf{x}}_{m}$ represents a clean or a distorted image. The trainable parameters in Equation (\ref{equ:target_loss}) are  ${\bf{W}}_{1}$, while all other network parameters are fixed and kept the same as those in the pre-trained models. Although Equation (\ref{equ:target_loss}) shows \emph{correction unit} estimation for only the first layer, it is possible to add such units at the output of distortion susceptible filters in any layer and in one or more layers. \figurename~\ref{fig:deep_corr} shows an example of a \emph{DeepCorrect} model for the pre-trained AlexNet model in \figurename~\ref{fig:AlexNet}.

\footnotetext[5]{The {\emph{correction unit}} is applied before the ReLU non-linearity acting upon the distortion-susceptible convolutional filter outputs. Max pooling layers following convolutional layers 1, 2 and 5 in the pre-trained AlexNet have not been shown in the ImageNet \emph{DeepCorrect} model for uniformity.}

\subsection{Rank-constrained DeepCorrect models}
\label{subsec:pruned} 
The inference time of our \emph{DeepCorrect} models (Section \ref{subsec:corr_unit}) can be slower than the respective baseline models due to the additional computational cost introduced by the \emph{correction units}. To mitigate the impact of these additional computations, we propose the use of a rank-constrained approximation to the full-rank \emph{DeepCorrect} model, which not only has the same computational cost as the corresponding baseline DNN but also retains almost 99\% of the performance of our full-rank \emph{DeepCorrect} models. 

Consider the $n^{th}$ full-rank 3-D convolutional filter with weights ${\bf{W}}_n \in \mathbb{R}^{k\times k \times C}$ in a DNN convolutional layer with $N$ filters, where $k\times k$ represents the filter's spatial extent  and $C$ is the number of input channels for the filter; then a rank constrained convolution is implemented by factorizing the convolution of ${\bf{W}}_n$ with input $z$ into a sequence of separable convolutions (i.e., horizontal and vertical filters) as in \cite{JaderbergVZ14}:
\begin{equation}
\small
{\bf{W}}_n*z \approx \sum\limits_{p=1}^{P}h_n^p*(v_p*z) =  \sum\limits_{p=1}^{P}h_n^p*\sum\limits_{c=1}^{C}v_p^c*z^c
\label{eqn:rank_const}
\end{equation}
\normalsize
where the first convolutional filter bank consists of $P$ vertical filters $\{v_p \in \mathbb{R}^{k\times 1 \times C}: p \in [1 ... P]\}$ and the second convolution consists of a horizontal filter that operates on $P$ input feature maps $\{h_n \in \mathbb{R}^{1\times k \times P} \}$. The number of intermediate filter maps, $P$, controls the rank of the low-rank approximation. The computational cost of the original full-rank convolutional layer for $N$ output feature maps with width $W'$ and height $H'$ is $O(Nk^2CH'W')$, whereas the rank-constrained approximation has a computational cost of $O((N+C)kPH'W')$ and a speedup can be achieved when $NkC > (N + C)P$. For the special case of convolutional layers in our \emph{correction units}, where $N = C$ (\figurename~\ref{fig:corr_unit}), if $P=N/2$, the computational cost of a convolutional layer can be reduced $k$ times (typically, $k = 3$). 

Our rank-constrained \emph{DeepCorrect} model is thus generated by replacing each full-rank convolutional layer (except $1 \times 1$ layers) with its respective rank-constrained approximation with $P$ for each approximation chosen such that the total computational cost of our model is the same as the baseline DNN. Instead of using iterative methods or training the separable filters from random weights, we use the simple yet fast, matrix decomposition approach of Tai \emph{et. al.} \cite{pruning} to get the exact global optimizer of the rank-constrained approximation from its respective trained full-rank \emph{DeepCorrect} model.

\section{Experimental Results}
\label{sec:exp}
We evaluate \emph{DeepCorrect} models against the alternative approaches of network fine-tuning and \emph{stability training} \cite{stability}, for the DNN architectures mentioned in Section \ref{sec:baseline}. The DNN models are trained and tested using a single Nvidia Titan-X GPU. Unlike common image denoising and deblurring methods like BM3D \cite{bm3d} and NCSR \cite{ncsr} which expect the distortion level to be known during both train and test phases or learning-based methods that train separate models for each distortion level, \emph{DeepCorrect} trains a single model for all distortion levels at once and, consequently, there is no need to know the distortion level at test time. 

\subsection{AlexNet Analysis}
\label{subsec:imagenet}

\subsubsection{Finetune model}
\label{imagenet_finetune}
 We fine-tune the AlexNet model in \figurename~\ref{fig:AlexNet} on a mix of distortion affected images and clean images to generate a single fine-tuned model and refer to this model as Finetune in our results. Starting with an initial learning rate (= 0.001) that is 10 times lower than that used to generate the pre-trained model in \figurename~\ref{fig:AlexNet} a fixed number of iterations (62500 iterations $\approx$ 10 epochs), with the learning rate reduced by a factor of 10 after every 18750 iterations (roughly 3 epochs). We use a data augmentation method as proposed by \cite{he2016deep}. 
 
 \begin{table}[]
\footnotesize
\renewcommand{\arraystretch}{1.4}
\caption{Top-1 accuracy of AlexNet-based DNN models for distortion affected images of the ImageNet validation set (ILSVRC2012), averaged over all levels of distortion and clean images. Bold numbers show best accuracy and underlined numbers show next best accuracy.}
\label{imagenet_blur_res}            
\centering
\begin{tabular}{|c|c|c|}
\hline
{\textbf{Method}} & {\textbf{Gaussian blur}} & {\textbf{AWGN}}\\\hline
Baseline & 0.2305 & 0.2375 \\

Finetune & 0.4596 & 0.4894 \\

Finetune-rc &0.4549 &0.4821 \\

\emph{Deepcorr} & {\textbf{0.5071}} & {\textbf{0.5092}}\\


\emph{Deepcorr}-b & \underline{0.5022} &\underline{0.5063}\\

\emph{Deepcorr}-rc &0.4992 &0.5052 \\

Stability\cite{stability} & 0.2163 &0.2305\\

NCSR\cite{ncsr}+AlexNet\cite{AlexNet} & 0.2193 & -\\
BM3D\cite{bm3d}+AlexNet\cite{AlexNet} & -	&0.5032 \\
\hline
\end{tabular}
\vspace*{-3ex}
\end{table}
\normalsize
\begin{table}[!t]
\scriptsize
\renewcommand{\arraystretch}{1.2}
\caption{Computational performance of AlexNet-based DNN models.}
\vspace*{-1ex}
\label{imagenet_blur_lite_res}            
\centering
\begin{tabular}{|c|c|c|c|c|}
\hline
{\textbf{Metric}}& {\textbf{Baseline/ Finetune}} & \textbf{\emph{Deepcorr}} &\textbf{\emph{Deepcorr}-b} &\textbf{\emph{Deepcorr}-rc}\\ \hline
FLOPs  & 7.4$\times10^8$
 &23.9$\times10^8$   & 11.8$\times10^8$ &7.8$\times10^8$\vspace{-2pt}\\
\begin{tabular}{@{}c@{}}
     Trainable      \vspace*{-4pt}\\
     params
\end{tabular}&60.96M & 2.81M& 1.03M & 1.03M \footnotemark[6]\\
\hline
\end{tabular}
\vspace*{-3ex}
\end{table}

\normalsize
 
 \subsubsection{Stability trained model}
 \label{stability}
 Following the \emph{stability training} method outlined in \cite{stability} that considers that unseen distortions can be modelled by adding AWGN to the input image, we fine-tune all the fully connected layers of the pre-trained AlexNet model by minimizing the KL-divergence between the classification scores for a pair of images ($I, I'$), where $I'$ = $I + \eta$ and $\eta \sim \mathcal{N}(0,\sigma^2)$. We use the same hyper-parameters used for the classification task in \cite{stability}: $\sigma^2$ = 0.04 and regularization coefficient $\alpha$ = 0.01.


\subsubsection{DeepCorrect models}
\label{imagenet_dc}
Our main \emph{DeepCorrect} model for AlexNet shown in \figurename~\ref{fig:AlexNet_corr} and referred to as \emph{Deepcorr} in Table \ref{imagenet_blur_res} is generated by correcting 75\% ranked filter outputs in the first two layers ($\beta_1, \beta_2$ = 0.75) and 50\% ranked filter outputs in the next three layers ($\beta_3, \beta_4, \beta_5$ = 0.5) of the pre-trained AlexNet shown in \figurename~\ref{fig:AlexNet}. The \emph{correction units} (\figurename~\ref{fig:corr_unit}) in each convolutional layer are trained using an initial learning rate of 0.1 and the same learning rate schedule, data augmentation and total iterations used for generating the Finetune model in Section \ref{imagenet_finetune}. We also generate two additional variants based on our \emph{Deepcorr} model, \emph{Deepcorr}-b, a computationally lighter model than \emph{Deepcorr} based on a bottleneck architecture for its \emph{correction units} as described later in this section, and \emph{Deepcorr}-rc, which is a rank-constrained model (Section~\ref{subsec:pruned}) derived from the full-rank \emph{Deepcorr}-b model such that its test-time computational cost is almost the same as the Finetune model. For comparison, a rank-constrained model is also derived for the Finetune model using similar decomposition parameters as \emph{Deepcorr}-rc and the resulting model is denoted by Finetune-rc.

\protect\footnotetext[6]{Since \emph{Deepcorr}-rc is derived from the \emph{Deepcorr}-b model through a low-rank approximation, the number of trainable parameters and subsequently its effect on model convergence is the same as \emph{Deepcorr}-b.}  

Table~\ref{imagenet_blur_res} shows the superior performance of our proposed method as compared to the alternative approaches and Table~\ref{imagenet_blur_lite_res} summarizes the computational performance in terms of trainable parameters and floating point operations (FLOPs) of these DNN models during training and testing, respectively. We evaluate the training computational cost in terms of the number of trainable parameters that are updated during training. The test-time computational cost is evaluated in terms of the total FLOPs, i.e., total multiply-add operations needed for a single image inference, as outlined by He \emph{et. al.} in \cite{he2016deep}. In particular, a $k\times k$ convolutional layer operating on $C$ input maps and producing $N$ output feature maps of width $W'$ and height $H'$ requires $Nk^2CH'W'$ FLOPs \cite{he2016deep}. The detailed architecture and corresponding FLOPs for each \emph{correction unit} in our different \emph{DeepCorrect} models for AlexNet are summarized in Table \ref{corr_arch}. Design choices for various \emph{correction unit} architectures and their impact on inference-time computational cost are also discussed later in this section.

As it can be seen from Table~\ref{imagenet_blur_res}, the \emph{Deepcorr} model, which is our best performing model in terms of top-1 classification accuracy, outperforms the Finetune model with $\approx$ 10\%  and $\approx$ 4\% relative average improvement for Gaussian blur and AWGN, respectively, by just training $\approx$ 2.81M parameters (Table \ref{imagenet_blur_lite_res}) as compared to 61M parameters for the Finetune model (i.e., 95.4\% lesser parameters). \emph{Deepcorr} significantly improves the robustness of a pre-trained DNN achieving an average top-1 accuracy of 0.5071 and 0.5092 for Gaussian blur and AWGN affected images, respectively, as compared to the corresponding top-1 accuracy of 0.2305 and 0.2375 for the pre-trained AlexNet DNN.

\begin{table*}[!t]
\renewcommand{\arraystretch}{1.3}
\centering
\caption{Correction unit architectures for AlexNet-based DeepCorrect models. The number following Corr-unit specifies the layer at which correction units are applied. Correction unit convolutional layers are represented as $k$x$k$, $d$, where $k$x$k$ is the spatial extent of a filter and $d$ is the number of filters in a layer. Stacked convolutional layers are enclosed in brackets, followed by the number of layers stacked.}
\vspace{-4pt}
\resizebox{0.72\textwidth}{!}{
\begin{threeparttable}
\begin{tabular}{c|c|c|c|c|c}
\hline
\scriptsize
\multirow{4}{*}{\textbf{Model}}   &   \multicolumn{1}{c|}{\textbf{Corr-unit 1}}   &   \multicolumn{1}{c|}{\textbf{Corr-unit 2}} &\multicolumn{1}{c|}{\textbf{Corr-unit 3}} &  \multicolumn{1}{c|}{\textbf{Corr-unit 4}}  & \multicolumn{1}{c} {\textbf{Corr-unit 5}}   \\
 &   \multicolumn{1}{c|}{ $\beta_1N_1$ = 72}   &   \multicolumn{1}{c|}{$\beta_2N_2$ = 192} & \multicolumn{1}{c|} {$\beta_3N_3$ = 192} &       \multicolumn{1}{c|}{$\beta_4N_4$ = 192}   & \multicolumn{1}{c} {$\beta_5N_5$ = 128}    \\
 &   \multicolumn{1}{c|}{\textbf{Output size}}   &   \multicolumn{1}{c|}{\textbf{Output size}} &\multicolumn{1}{c|}{\textbf{Output size}} &  \multicolumn{1}{c|}{\textbf{Output size}}  & \multicolumn{1}{c} {\textbf{Output size}}   \\
 
  &   \multicolumn{1}{c|}{$55\times55$}   &   \multicolumn{1}{c|}{$27\times27$} &\multicolumn{1}{c|}{$13\times13$} &  \multicolumn{1}{c|}{$13\times13$}  & \multicolumn{1}{c} {$13\times13$}   \\
 \Xhline{3\arrayrulewidth}
\multirow{3}{*}{\emph{Deepcorr}} & 
      1$\times$1, 72   & 1$\times$1, 192 &1$\times$1, 192  & 1$\times$1, 192 & 1$\times$1, 128 \\
                                                                                   &\hspace{2ex} 
                  $\begin{bmatrix}\text{ 5$\times$5, 72 }\end{bmatrix}$$\times$2                 
      & \hspace{2ex} 
      $\begin{bmatrix}\text{ 3$\times$3, 192 }\end{bmatrix}$$\times$2                 &\hspace{2ex}  $\begin{bmatrix}\text{ 3$\times$3, 192 }\end{bmatrix}$$\times$2                  &\hspace{2ex}  $\begin{bmatrix} \text{ 3$\times$3, 192 }\end{bmatrix}$$\times$2                &\hspace{2ex}  
      $\begin{bmatrix} \text{ 3$\times$3, 128 }\end{bmatrix}$$\times$2                            \\
                  &   1$\times$1, 72                &  1$\times$1, 192                 &   1$\times$1, 192                 &1$\times$1, 192                   &    1$\times$1, 128 
                \\
                  \multirow{1}{*}{Total FLOPs: 16.5$\times10^8$ }   
                 & FLOPs: 8.1$\times10^8$   &  FLOPs: 5.3$\times10^8$  &   FLOPs: 1.2$\times10^8$      & FLOPs: 1.2$\times10^8$             & FLOPs: 5.5$\times10^7$ 
                  \\\hline
\multirow{3}{*}{\emph{Deepcorr}-b} &1$\times$1, 36 &1$\times$1, 96  &1$\times$1, 96  &1$\times$1, 96  &1$\times$1, 64  \\
  &\hspace{2ex}  
  $\begin{bmatrix}\text{ 3$\times$3, 36 }\end{bmatrix}$$\times$3 &\hspace{2ex} 
  $\begin{bmatrix} \text{ 3$\times$3, 96 }\end{bmatrix}$$\times$3                &\hspace{2ex} $\begin{bmatrix}\text{ 3$\times$3, 96 }\end{bmatrix}$$\times$3 &\hspace{2ex} 
  $\begin{bmatrix} \text{ 3$\times$3, 96 }\end{bmatrix}$$\times$3 &\hspace{2ex}  
  $\begin{bmatrix} \text{ 3$\times$3, 64 }\end{bmatrix}$$\times$3                             \\
  &   1$\times$1, 72                & 1$\times$1, 192                  &  1$\times$1, 192                 & 1$\times$1, 192                  &  1$\times$1, 128   
\\\multirow{1}{*}{Total FLOPs: 4.4$\times10^8$} 
& FLOPs: 1.2$\times10^8$    &FLOPs: 2.0$\times10^8$     &FLOPs: 4.8$\times10^7$     &FLOPs: 4.8$\times10^7$  &FLOPs: 2.1$\times10^7$
  \\\hline         
  \multirow{4}{*}{\emph{Deepcorr}-rc} &1$\times$1, 36 &1$\times$1, 96  &1$\times$1, 96  &1$\times$1, 96  &1$\times$1, 64  \\
  &\hspace{2ex}  
  $\begin{bmatrix}\text{ 3$\times$1, 27 }\\\text{ 1$\times$3, 36 }\end{bmatrix}$$\times$3 &\hspace{2ex} 
  $\begin{bmatrix} \text{ 3$\times$1, 72 }\\\text{1$\times$3, 96}\end{bmatrix}$$\times$3              & \hspace{2ex} $\begin{bmatrix}\text{ 3$\times$1, 72 }\\
  \text{1$\times$3, 96}\end{bmatrix}$$\times$3 &\hspace{2ex} 
  $\begin{bmatrix} \text{ 3$\times$1, 72 }\\\text{ 1$\times$3, 96 }\end{bmatrix}$$\times$3 & \hspace{2ex}
  $\begin{bmatrix} \text{ 3$\times$1, 48 }\\\text{ 1$\times$3, 64 }\end{bmatrix}$$\times$3                             \\
  &   1$\times$1, 72                & 1$\times$1, 192                  &  1$\times$1, 192                 & 1$\times$1, 192                  &  1$\times$1, 128
\\\multirow{1}{*}{Total FLOPs: 2.5$\times10^8$} 
&FLOPs: 6.8$\times10^7$     &FLOPs: 1.1$\times10^8$     &FLOPs: 2.7$\times10^7$      &FLOPs: 2.7$\times10^7$    &FLOPs: 1.2$\times10^7$ 
  \\\hline   
\end{tabular} 
\vspace*{-3ex}
\end{threeparttable}}
\label{corr_arch}
\end{table*}

All our \emph{DeepCorrect} model variants consistently outperform the Finetune and Stability models for both distortion types (Table \ref{imagenet_blur_res}). We also observe that fine-tuning DNN models on distortion specific data significantly outperforms DNN models trained through distortion agnostic \emph{stability training}. For completeness, we also compare classification performance with AlexNet when combined with a commonly used non-blind image denoising method (BM3D) proposed by \cite{bm3d} and a deblurring method (NCSR) proposed by \cite{ncsr}, where BM3D and NCSR are applied prior to the baseline AlexNet, for AWGN and Gaussian blur, respectively. 
Table~\ref{imagenet_blur_res} shows top-1 accuracy for these two methods with 
the \emph{Deepcorr} model outperforming each for AWGN and Gaussian blur, respectively. 

\captionsetup[subfigure]{labelformat=empty}
\begin{figure*}[]
\centering
\subfloat[]{\includegraphics[width=0.9\textwidth]{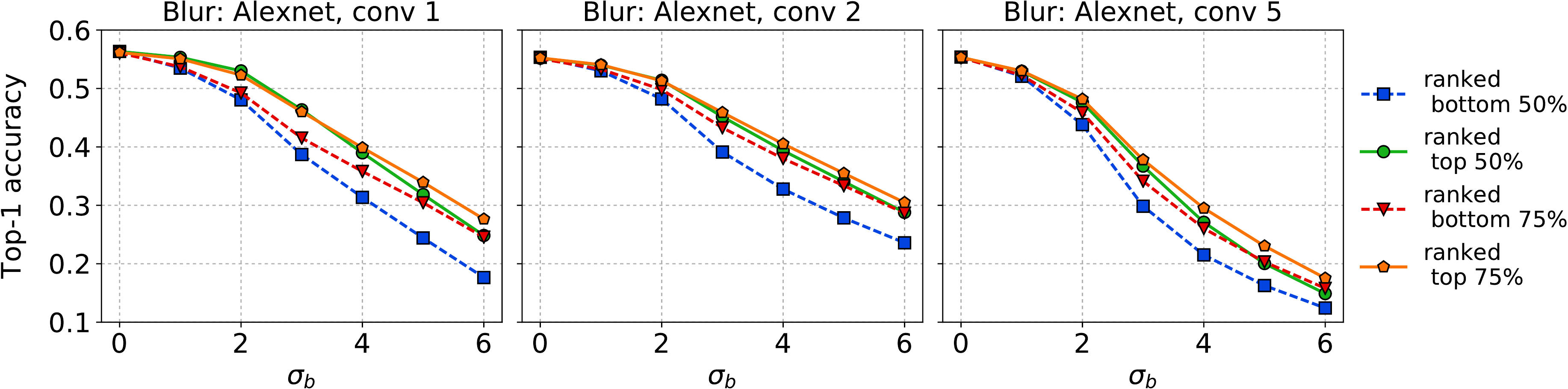}
\label{fig:blur_rank}
}
\vspace{-8pt}
\subfloat[]{\includegraphics[width=0.9\textwidth]{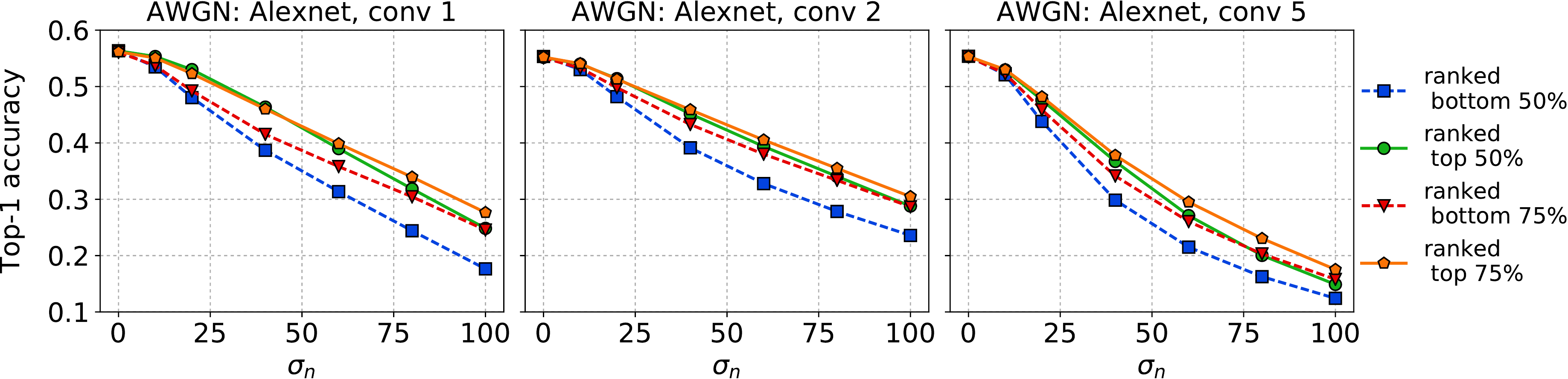}
\label{fig:awgn_rank}
}
\vspace*{-12pt}
\caption[]{Effect of \emph{DeepCorrect} ranking metric on \emph{correction unit} performance when integerated with AlexNet\cite{AlexNet}. Dashed lines represent \emph{correction units} trained on the least susceptible filters in a DNN layer and solid lines represent \emph{correction units} trained on the most susceptible filters, as identified by our ranking metric (Section \ref{subsec:ranking}).}
\label{fig:ranking_exp}
\vspace*{-2ex}
\end{figure*}

\subsubsection{Correction unit architectures}
\label{subsec:imagenet_arch}
Increasing the correction unit kernel depth ($D_i$ in \figurename~\ref{fig:corr_unit}) makes our proposed \emph{correction unit} fatter, whereas decreasing $D_i$ makes the \emph{correction unit} thinner. A natural choice for $D_i$ would be to make it equal to the number of distortion susceptible filters that need correction ($\beta_iN_i$), in a DNN layer $i$; this is also the default parameter setting used in the \emph{correction units} for \emph{Deepcorr}. Since $D_i$ is always equal to the number of distortion susceptible filters that need correction (i.e., $D_i$ = $\beta_iN_i$), the number of trainable parameters in the \emph{correction units} of \emph{Deepcorr} scale linearly with the number of corrected filters ($\beta_iN_i$) in each convolutional layer, even though \emph{Deepcorr} trains significantly lesser parameters than Finetune and still achieves a better classification accuracy (Tables \ref{imagenet_blur_res} and \ref{imagenet_blur_lite_res}). 
As shown in Table~\ref{imagenet_blur_lite_res}, during the testing phase, the \emph{correction units} in \emph{Deepcorr} add almost 2 times more FLOPs relative to the baseline DNN, for evaluating a single image.
As discussed in more detail later in this section, one way to limit the number of trainable parameters and FLOPs is to explore a bottleneck architecture for our \emph{correction units}, where the convolutional kernel depth $D_i$ (\figurename~\ref{fig:corr_unit}) is set to 50\% of the distortion susceptible filters that need correction in a DNN layer $i$ (i.e., $D_i=\frac{\beta_iN_i}{2}$) as compared to $D_i$ being set to the number of filters to correct (i.e., $D_i = \beta_iN_i$) in \emph{Deepcorr}. Replacing each \emph{correction unit} in \emph{Deepcorr} with such a bottleneck \emph{correction unit}, which we refer to as \emph{Deepcorr}-b, results in a \emph{DeepCorrect} model that has significantly less trainable parameters and FLOPs than the original \emph{Deepcorr} model (Table \ref{corr_arch}). Compared to the \emph{correction units} in \emph{Deepcorr}, bottleneck \emph{correction units} provide a 60\% reduction in FLOPs on average, with a 85\% reduction in FLOPs for Corr-unit 1 as shown in Table~\ref{corr_arch}. 
From Tables \ref{imagenet_blur_res} and \ref{imagenet_blur_lite_res}, it can be seen that \emph{Deepcorr}-b achieves $\approx$ 99\% of the average accuracy achieved by \emph{Deepcorr}, with a 63\% reduction in trainable parameters and a 73\% reduction in FLOPs, relative to \emph{Deepcorr}. 
As shown in Table~\ref{imagenet_blur_lite_res}, the \emph{Deepcorr}-b model still requires 58\% more FLOPs relative to the baseline DNN, for evaluating a single image at test time. A further reduction in the computational cost at test-time can be achieved by deriving a rank-constrained \emph{DeepCorrect} model (\emph{Deepcorr}-rc) from the \emph{Deepcorr}-b model using the approach outlined in Section~\ref{subsec:pruned}. By replacing each full rank convolutional layer in a \emph{Deepcorr}-b \emph{correction unit} with a pair of separable convolutions (Section~\ref{subsec:pruned}), we can reduce the FLOPs for each \emph{correction unit} by an additional 46\% as shown in Table~\ref{corr_arch}. Replacing the pre-trained convolutional layers of the baseline AlexNet (which are left unchanged in the \emph{DeepCorrect} models and shown in gray in \figurename~\ref{fig:deep_corr}) by their equivalent low-rank approximations provides an additional 29\% reduction in FLOPs, relative to the baseline AlexNet model, such that the resultant \emph{Deepcorr}-rc model now has almost the same FLOPs as Finetune (Table~\ref{imagenet_blur_lite_res}) and still retains almost 99\% of the accuracy achieved by \emph{Deepcorr}.

\subsubsection{Effect of ranking on correction unit performance}
If the superior performance of our \emph{DeepCorrect} model is only due to the additional network parameters provided by the \emph{correction unit}, then re-training the correction unit on the least susceptible $\beta_iN_i$ filter outputs in a DNN layer should also achieve the same performance as that achieved by applying a \emph{correction unit} on the $\beta_iN_i$ most susceptible filter outputs identified by our ranking metric (Section \ref{subsec:ranking}). $\beta_i$ and $N_i$, as defined in Section \ref{subsec:ranking}, represent the percentage of filters corrected in the $i^{th}$ layer and the total number of filters in the $i^{th}$ layer, respectively. To this end, we evaluate the performance of training \emph{correction units} on: 1) $\beta_iN_i$ filters most susceptible to distortion, and 2) $\beta_iN_i$ filters least susceptible to distortion, as identified by our ranking metric. For this analysis, $\beta_i \in $ \{50\%, 75\%\}. 

As shown in \figurename~\ref{fig:ranking_exp}, \emph{correction units} trained on the $\beta_iN_i$ most distortion susceptible filters (solid lines) outperform those trained on the least susceptible filters (dashed lines) of conv 1, conv 2 and conv 5 layers of the AlexNet model, for Gaussian blur and AWGN, respectively. Although we observe a similar trend for the conv 3 and conv 4 layers, we plot results for only conv 1, conv 2 and conv 5 as these show the largest difference in performance due to ranking distortion susceptible filters. Correcting the top 75\% distortion susceptible filters (solid orange), as identified by our ranking measure, achieves the best performance in all layers. Similarly, for all layers and distortions, \emph{correction units} trained on the top 50\% susceptible filters (solid green) not only significantly outperform \emph{correction units} trained on the 50\% least susceptible filters (dashed blue) but also outperform \emph{correction units} trained on the 75\% least susceptible filters (dashed red), where 25\% filters are shared among both approaches.

\subsubsection{Accelerating training}
\begin{figure}[]
\centering
\includegraphics[width=\linewidth]{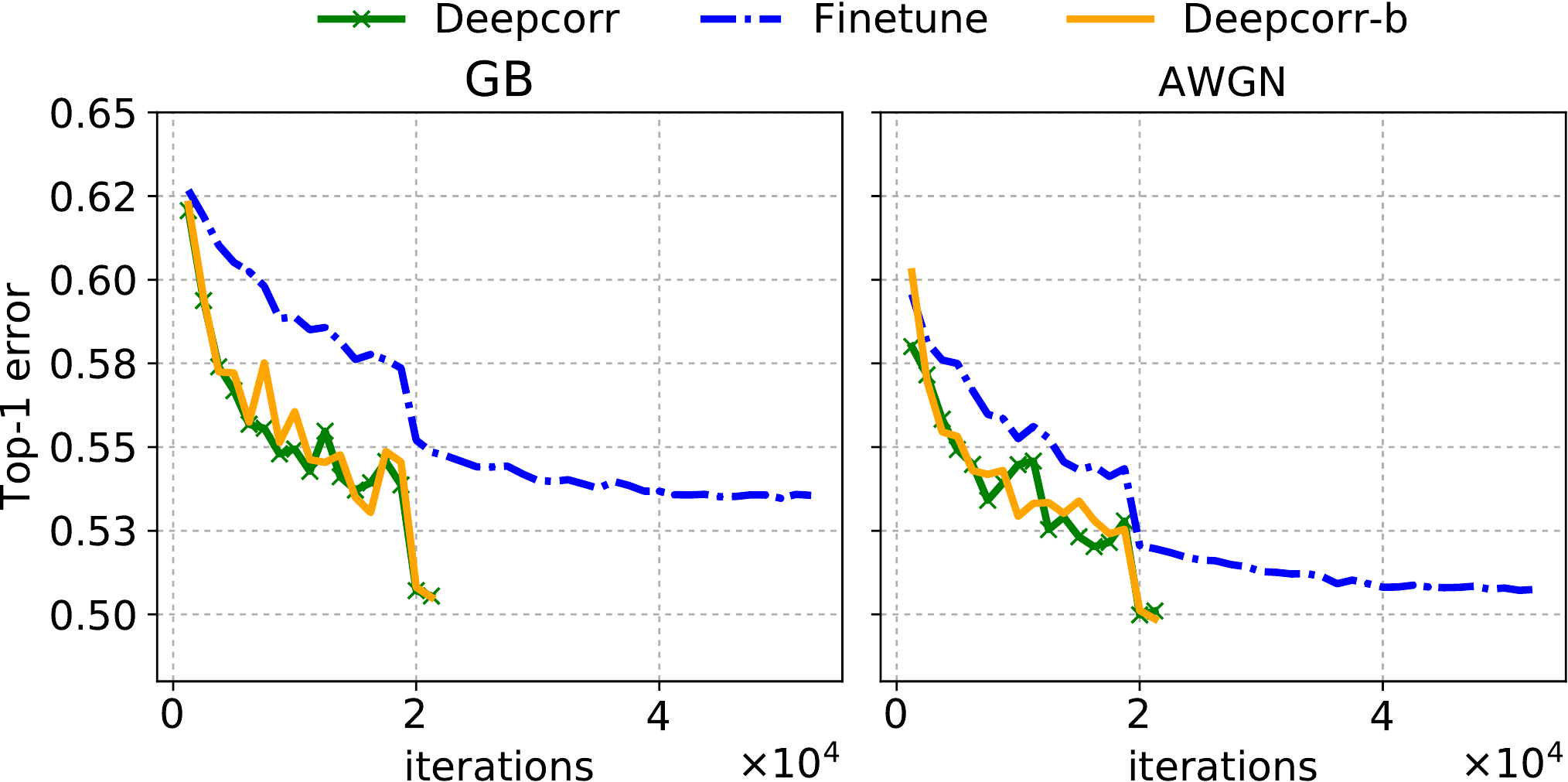}
\vspace{-16pt}
\caption{Top-1 error on the ILSVRC-2012 validation set, for \emph{DeepCorrect} model variants and fine-tuning. 
\vspace{-16pt}
} 
\label{fig:training_time}
\end{figure}

We analyze the evolution of the validation set error over training iterations for \emph{Deepcorr}, \emph{Deepcorr}-b as well as the Finetune model and stop training for the \emph{Deepcorr} models when their respective validation error is lesser than or equal to the minimum validation error achieved by the Finetune model. 
For any particular value of validation error achieved by the Finetune model in \figurename~\ref{fig:training_time}, both the \emph{Deepcorr} model variants are able to achieve the same validation error in much lesser number of training iterations. Thus, we conclude that just learning corrective transforms for activations of a subset of convolutional filters in a layer accelerates training through reduced number of training epochs needed for convergence.
\subsubsection{Generalization of DeepCorrect Features to other Datasets}
\label{subsec:caltech}
To analyze the ability of distortion invariant features learnt for image classification to generalize to  related tasks like object recognition and scene recognition, we evaluate the \emph{Deepcorr} and Finetune models trained on the ImageNet dataset (Section \ref{subsec:imagenet}) as discriminative deep feature extractors on the Caltech-101 \cite{caltech-101}, Caltech-256 \cite{caltech-256}  and SUN-397 \cite{sun397} datasets. 
Unlike object recognition datasets like Caltech-101 and Caltech-256, which bear some similarity to an image classification/object recognition dataset like ImageNet, a scene recognition dataset like SUN-397 bears no similarity to the ImageNet dataset and is expected to be challenging for features extracted using models learnt on ImageNet \cite{decaf}. 
Following the experimental procedure proposed by \cite{decaf}, we use the output of the first (Caltech-101 and Caltech-256) or second (SUN-397) fully-connected layer in these models as a deep feature extractor for images affected by distortion. 

Since the above deep feature models have not been trained on any one of these datasets (Caltech-101, Caltech-256 and SUN-397), for each dataset, we train linear SVMs on top of the deep features, which are extracted from a random set of training data, and evaluate the performance in terms of mean accuracy per category averaged over 5 data splits, following the training procedure adopted by \cite{decaf}. The training data for each split consists of 25 training images and 5 validation images per class, sampled randomly from the considered dataset and all remaining images are used for testing. A baseline accuracy for undistorted images is first established by training linear SVMs on features extracted only from undistorted images using the AlexNet DNN shown in \figurename~\ref{fig:AlexNet}, and results are reported in Table \ref{baseline_decaf}. 
\begin{table}[]
\scriptsize
\renewcommand{\arraystretch}{1.3}
\setlength{\tabcolsep}{1.5em}
\caption{Mean accuracy per category of pre-trained AlexNet deep feature extractor for clean images. }
\vspace*{-1ex}
\label{baseline_decaf}            
\centering
\begin{tabular}{|c|c|c|}
\hline
Caltech-101 & Caltech-256 & SUN-397\\
\Xhline{3\arrayrulewidth}
0.8500 &0.6200 &0.3100
\\\hline
\end{tabular}
\vspace*{-3ex}
\end{table}
\begin{table}[]
\scriptsize
\renewcommand{\arraystretch}{1.3}
\setlength{\tabcolsep}{1.25em}
\caption{Mean accuracy per category for Gaussian blur affected images, averaged over all distortion levels. Bold numbers show best accuracy. }
\vspace*{-1ex}
\label{caltech_blur_res}            
\centering
\begin{tabular}{|c|c|c|c|}
\hline
Dataset & Baseline & Finetune & \emph{Deepcorr}\\
\Xhline{3\arrayrulewidth}
Caltech-101 & 0.4980  &0.7710 &\textbf{0.8371} \\
Caltech-256 & 0.2971  &0.5167 &\textbf{0.5883} \\
SUN-397 &0.1393  &0.2369 &\textbf{0.3049} 
\\\hline
\end{tabular}
\vspace*{-3ex}
\end{table}
\begin{table}[!t]
\scriptsize
\renewcommand{\arraystretch}{1.3}
\setlength{\tabcolsep}{1.25em}
\caption{Mean accuracy per category for AWGN affected images, averaged over all distortion levels. Bold numbers show best accuracy. }
\vspace*{-1ex}
\label{caltech_awgn_res}            
\centering
\begin{tabular}{|c|c|c|c|}
\hline
Dataset & Baseline & Finetune & \emph{Deepcorr}\\
\Xhline{3\arrayrulewidth}
Caltech-101 & 0.3423  &0.7705 &\textbf{0.8034} \\
Caltech-256 & 0.1756  & 0.4995 & \textbf{0.5482} \\
SUN-397 &0.0859  &0.1617 &\textbf{0.2936}
\\\hline
\end{tabular}
\vspace*{-4ex}
\end{table}
\normalsize
Similar to the evaluation in Section \ref{subsec:imagenet}, we now independently add Gaussian blur and AWGN to train and test images using the same distortion levels as reported in Section \ref{sec:baseline} and report performance averaged over all distortion levels in Tables\ref{caltech_blur_res}-\ref{caltech_awgn_res} for deep features extracted using baseline AlexNet, Finetune and \emph{Deepcorr} models trained on ImageNet\footnote[7]{For each of the three models, a single set of linear SVMs is trained for images affected by different levels of distortion and also clean images.}. 
Both Gaussian blur and AWGN significantly affect the accuracy of the baseline feature extractor for all three datasets, with a 41\% and 60\% drop in respective accuracies for Caltech-101, a 52\% and 71\% drop in respective accuracies for Caltech-256, and a 55\% and 72\% drop in respective mean accuracy for SUN-397, relative to the benchmark performance for clean images. For Caltech-101, the \emph{Deepcorr} feature extractor outperforms the Finetune feature extractor with a 8.5\% and 4.2\% relative improvement in mean accuracy for Gaussian blur and AWGN affected images, respectively. For Caltech-256, the \emph{Deepcorr} feature extractor outperforms the Finetune feature extractor with a 13.8\% and 9.7\% relative improvement for Gaussian blur and AWGN, respectively. Similarly, features extracted using the \emph{Deepcorr} model significantly outperform those extracted using the Finetune model for SUN-397, with a 28.7\% and 81.5\% relative improvement in mean accuracy for Gaussian blur and AWGN, respectively. The large performance gap between \emph{Deepcorr} and Finetune feature extractors highlights the generic nature of distortion invariant features learnt by our \emph{DeepCorrect} models.
\vspace*{-1ex}
\begin{table}[]
\footnotesize
\renewcommand{\arraystretch}{1.2}
\caption{Top-1 accuracy of ResNet18-based DNN models for distortion affected images of the ImageNet validation set (ilsvrc2012), averaged over all levels of distortion and clean images. Bold numbers show best accuracy and underlined numbers show next best accuracy.}
\vspace*{-1ex}
\label{res18_blur_res}            
\centering
\begin{tabular}{|c|c|c|c|c|c|}
\hline
{\textbf{Method}} & {\textbf{G.Blur}} & {\textbf{AWGN}} & {\textbf{M.Blur}} & {\textbf{D.Blur}}
& {\textbf{Cam.Blur}}\\\hline
Baseline &0.3841 & 0.3255 &0.4436 &0.3582 &0.4749\\
Finetune &0.5617 & 0.5970 &0.6197 &0.5615 &0.6041 \\
Finetune-rc &0.5548 &0.5898 &0.6113 &0.5537 &0.5973 \\
\emph{Deepcorr} &0.5808 &\underline{0.6058} &{\bf{0.6498}} &{\bf{0.6005}} &\underline{0.6346} \\
\emph{Deepcorr}-b &{\bf{0.5839}} &{\bf{0.6087}} &\underline{0.6474} &\underline{0.5831} &{\bf{0.6365}} \\
\emph{Deepcorr}-rc &\underline{0.5821} &0.6033 &0.6411 &0.5785 &0.6276 \\
Stability\cite{stability} &0.3412 &0.3454 &0.4265 &0.3182 &0.4720  \\
\hline
\end{tabular}
\vspace*{-2ex}
\end{table}
\begin{table}[]
\renewcommand{\arraystretch}{1.2}
\caption{Computational performance of ResNet18-based dnn models.}
\vspace*{-1ex}
\scriptsize
\label{res18_flops}            
\centering
\begin{threeparttable}
\begin{tabular}{|c|c|c|c|c|}
\hline
{\textbf{Metric}} &{\textbf{Baseline/ Finetune}} & {\textbf{\emph{Deepcorr}}} & {\textbf{\emph{Deepcorr}-b}} &{\textbf{\emph{Deepcorr}-rc}}\\\hline
FLOPs  &1.8$\times10^9$ &3.5$\times10^9$ &2.9$\times10^9$ &1.8$\times10^9$ \vspace{-2pt}\\
\begin{tabular}{@{}c@{}}
     Trainable      \vspace*{-4pt}\\
     params
\end{tabular} &11.7M &8.41M &5.5M & 5.5M\\
\hline
\end{tabular}
\vspace*{-3ex}
\end{threeparttable}
\end{table}
\subsection{ResNet18 Analysis}
\label{subsec:res18}
\label{subsec:res18}
Similar to the AlexNet analysis in Section \ref{subsec:imagenet}, for ResNet18, we evaluate the performance of our proposed \emph{DeepCorrect} models against DNN models trained through fine-tuning and \emph{stability training}. In addition to Gaussian blur (GB) and AWGN, we also evaluate 3 additional distortion types used by Vasiljevic \emph{et. al.} \cite{fine_tune}, namely 1) Motion blur (M.Blur), 2) Defocus blur (D.Blur) and 3) Camera shake blur (Cam.Blur). Spatially uniform disk kernels of varying radii are used to simulate defocus blur while horizontal and vertical box kernels of single pixel width and varying length are used to simulate motion blur (uniform linear motion)\cite{fine_tune}. For simulating camera shake blur, we use the code in \cite{blur_kernels} to generate 100 random blur kernels and report performance averaged over all camera shake blur kernels.

Using the training procedures outlined in Sections~\ref{imagenet_finetune} and \ref{stability}, we generate fine-tuned and \emph{stability trained} models from the baseline ResNet18 DNN by training all layers (11.7M parameters) of the DNN on a mix of distorted and clean images. Similarly, using the training procedure of Section~\ref{imagenet_dc}, our competing \emph{DeepCorrect} model is generated by training \emph{correction units} that are appended at the output of the most susceptible filters in the odd-numbered convolutional layers (1 to 17) of the baseline ResNet18 model right after each skip connection merge (\figurename~\ref{fig:res18}). Similar to the AlexNet analysis presented earlier (Section~\ref{subsec:imagenet}), we generate three \emph{DeepCorrect models} (i.e., \emph{Deepcorr}, \emph{Deepcorr}-b and \emph{Deepcorr}-rc). For comparison, a rank-constrained model is derived for the Finetune model using similar decomposition parameters as \emph{Deepcorr}-rc and the resulting model is denoted by Finetune-rc. Table \ref{res18_blur_res} summarizes the accuracy of ResNet18-based DNN models against various distortions, while Table \ref{res18_flops} shows the computational performance of the same DNN models, measured in terms of FLOPs and trainable parameters. Similar to the AlexNet analysis, the detailed architecture and corresponding FLOPs for each \emph{correction unit} in our different \emph{DeepCorrect} models for ResNet18 are shown in Table \ref{corr_arch_resnet}.

\begin{table*}[!t]
\renewcommand{\arraystretch}{1.3}
\centering
\caption{Correction unit architectures for ResNet18-based DeepCorrect models. The number following Corr-unit specifies the layer at which correction units are applied. Correction unit convolutional layers are represented as $k$x$k$, $d$, where $k$x$k$ is the spatial extent of a filter and $d$ is the number of filters in a layer. Stacked convolutional layers are enclosed in brackets, followed by the number of layers stacked.}
\vspace{-4pt}
\resizebox{0.72\textwidth}{!}{
\begin{threeparttable}
\begin{tabular}{c|c|c|c|c|c}
\hline
\multirow{4}{*}{\textbf{Model}}   &   \multicolumn{1}{c|}{\textbf{Corr-unit 1}}   &   \multicolumn{1}{c|}{\textbf{Corr-unit 3, 5 }} &\multicolumn{1}{c|}{\textbf{Corr-unit 5, 7}} &  \multicolumn{1}{c|}{\textbf{Corr-unit 9, 11}}  & \multicolumn{1}{c} {\textbf{Corr-unit 13, 15}} \\
 &   \multicolumn{1}{c|}{ $\beta_1N_1$ = 48}   &   \multicolumn{1}{c|}{$\beta_2N_2$ = 48} & \multicolumn{1}{c|} {$\beta_3N_3$ = 96} &       \multicolumn{1}{c|}{$\beta_4N_4$ = 192}   & \multicolumn{1}{c} {$\beta_5N_5$ = 384}\\
 &   \multicolumn{1}{c|}{\textbf{Output size}}   &   \multicolumn{1}{c|}{\textbf{Output size}} &\multicolumn{1}{c|}{\textbf{Output size}} &  \multicolumn{1}{c|}{\textbf{Output size}}  & \multicolumn{1}{c} {\textbf{Output size}}   \\
 
  &   \multicolumn{1}{c|}{$112\times112$}   &   \multicolumn{1}{c|}{$56\times56$} &\multicolumn{1}{c|}{$28\times28$} &  \multicolumn{1}{c|}{$14\times14$}  & \multicolumn{1}{c} {$7\times7$}
 
 \\\Xhline{3\arrayrulewidth}
\multirow{3}{*}{\emph{Deepcorr}} & 
      1$\times$1, 48   & 1$\times$1, 48 &1$\times$1, 96  & 1$\times$1, 192 & 1$\times$1, 384 \\
      &\hspace{2ex} 
                  $\begin{bmatrix}\text{ 3$\times$3, 48 }\end{bmatrix}$$\times$2                 & \hspace{2ex} $\begin{bmatrix}\text{ 3$\times$3, 48 }\end{bmatrix}$$\times$2                 &\hspace{2ex}  $\begin{bmatrix}\text{ 3$\times$3, 96 }\end{bmatrix}$$\times$2                  &\hspace{2ex}  $\begin{bmatrix} \text{ 3$\times$3, 192 }\end{bmatrix}$$\times$2                &\hspace{2ex}  $\begin{bmatrix} \text{ 3$\times$3, 384 }\end{bmatrix}$$\times$2\\
                  &   1$\times$1, 48                &  1$\times$1, 48                 &   1$\times$1, 96                 &1$\times$1, 192                   &    1$\times$1, 384 \\
                  \multirow{1}{*}{Total FLOPs: 1.7$\times10^9$}   
                 &FLOPs: 5.7$\times10^8$   &FLOPs: 1.4$\times10^8$  &FLOPs: 1.4$\times10^8$      &FLOPs: 1.4$\times10^8$             &FLOPs: 1.4$\times10^8$ \\\hline
\multirow{3}{*}{\emph{Deepcorr}-b} &1$\times$1, 31 &1$\times$1, 31  &1$\times$1, 62  &1$\times$1, 124  &1$\times$1, 248 \\
  &\hspace{2ex}  
  $\begin{bmatrix}\text{ 3$\times$3, 31 }\end{bmatrix}$$\times$3 &\hspace{2ex} 
  $\begin{bmatrix} \text{ 3$\times$3, 31 }\end{bmatrix}$$\times$3                &\hspace{2ex} $\begin{bmatrix}\text{ 3$\times$3, 62 }\end{bmatrix}$$\times$3 &\hspace{2ex} 
  $\begin{bmatrix} \text{ 3$\times$3, 124 }\end{bmatrix}$$\times$3 &\hspace{2ex}  
  $\begin{bmatrix} \text{ 3$\times$3, 248 }\end{bmatrix}$$\times$3  \\
  &   1$\times$1, 48                & 1$\times$1, 48                  &  1$\times$1, 96                 & 1$\times$1, 192                  &  1$\times$1, 384   \\
\multirow{1}{*}{Total FLOPs: 1.1$\times10^9$ } 
&FLOPs: 3.6$\times10^8$    &FLOPs: 9.2$\times10^7$     &FLOPs: 9.2$\times10^7$     &FLOPs: 9.2$\times10^7$  &FLOPs: 9.2$\times10^7$ \\\hline         
  \multirow{4}{*}{\emph{Deepcorr}-rc} &1$\times$1, 31 &1$\times$1, 31  &1$\times$1, 62  &1$\times$1, 124  &1$\times$1, 248 \\
  &\hspace{2ex}  
  $\begin{bmatrix}\text{ 3$\times$1, 24 }\\\text{ 1$\times$3, 31 }\end{bmatrix}$$\times$3 
  &\hspace{2ex} 
  $\begin{bmatrix} \text{ 3$\times$1, 24 }\\\text{1$\times$3, 31}\end{bmatrix}$$\times$3              
  & \hspace{2ex} 
  $\begin{bmatrix}\text{ 3$\times$1, 48 }\\
  \text{ 1$\times$3, 62}\end{bmatrix}$$\times$3 
  &\hspace{2ex} 
  $\begin{bmatrix} \text{ 3$\times$1, 96 }\\\text{ 1$\times$3, 124  }\end{bmatrix}$$\times$3 
  & \hspace{2ex} 
  $\begin{bmatrix} \text{ 3$\times$1, 192 }\\\text{ 1$\times$3, 248 }\end{bmatrix}$$\times$3                            \\
  &   1$\times$1, 48                & 1$\times$1, 48                  &  1$\times$1, 96                 & 1$\times$1, 192                  &  1$\times$1, 384 \\
\multirow{1}{*}{Total FLOPs: 0.6$\times10^9$} 
&FLOPs: 2.0$\times10^8$     &FLOPs: 5.1$\times10^7$     &FLOPs: 5.1$\times10^7$      &FLOPs: 5.1$\times10^7$    &FLOPs: 5.1$\times10^7$ \\\hline   
\end{tabular} 
\vspace*{-4ex}
\end{threeparttable}}
\label{corr_arch_resnet}
\end{table*}

From Table \ref{res18_blur_res}, we observe that our \emph{DeepCorrect} models not only significantly improve the robustness of the baseline DNN model to input distortions but also outperform the alternative approaches of model fine-tuning and \emph{stability training} in terms of classification accuracy. \emph{Deepcorr}, which trains almost 28\% lesser parameters (Table~\ref{res18_flops}) than Finetune, outperforms Finetune by 3.2\% for Gaussian blur, 1.47\% for AWGN, 4.86\% for motion blur, 6.95\% for defocus blur and 5\% for camera shake blur. \emph{Deepcorr}-b, which trains 50\% lesser parameters (Table~\ref{res18_flops}) than Finetune, outperforms Finetune by 3.95\% for Gaussian blur, 1.9\% for AWGN, 4.46\% for motion blur, 3.84\% for defocus blur and 5.36\% for camera shake blur. Similarly our rank constrained model (\emph{Deepcorr}-rc), which trains 50\% parameters less than Finetune (Table~\ref{res18_flops}), outperforms Finetune by 3.63\% for Gaussian blur, 1\% for AWGN, 3.45\% for motion blur, 3\% for defocus blur and 3.89\% for camera shake blur. As shown in Tables~\ref{res18_flops} and~\ref{corr_arch_resnet}, during the testing phase, \emph{Deepcorr}-b requires 60\% more FLOPs relative to the baseline DNN, for evaluating a single image. On the other hand, for \emph{Deepcorr}-rc, just using rank-constrained \emph{correction units} results in a 50\% reduction in FLOPs as compared to \emph{Deepcorr}-b (Table~\ref{corr_arch_resnet}), with a 30\% additional reduction in FLOPs relative to baseline ResNet18 achieved by replacing the pre-trained convolutional layers of baseline ResNet18 with their low-rank approximations. From Tables \ref{res18_blur_res} and \ref{res18_flops}, it can be seen that \emph{Deepcorr}-rc requires almost the same number of FLOPs as Finetune but achieves a superior performance than the Finetune and Stability models without sacrificing inference speed. 

\section{Conclusion}
\label{sec:conc}
Deep networks trained on pristine images perform poorly when tested on distorted images affected by image blur or additive noise. Evaluating the effect of Gaussian blur and AWGN on the activations of convolutional filters trained on undistorted images, we observe that select filters in each DNN convolutional layer are more susceptible to input distortions than the rest. We propose a novel objective metric to assess the susceptibility of convolutional filters to distortion and use this metric to identify the filters that maximize DNN robustness to input distortions, upon correction of their activations. 

We design \emph{correction units}, which are \emph{residual blocks} that are comprised of a small stack of trainable convolutional layers and a single skip connection per stack. These \emph{correction units} are added at the output of the most distortion susceptible filters in each convolutional layer, whilst leaving the rest of the pre-trained (on undistorted images) filter outputs in the network unchanged. The resultant DNN models which we refer to as \emph{DeepCorrect} models, significantly improve the robustness of DNNs against image distortions and also outperform the alternative approach of network fine-tuning on common vision tasks like image classification, object recognition and scene classification, whilst training significantly less parameters and achieving a faster convergence in training. Fine-tuning limits the ability of the network to learn invariance to severe levels of distortion, and re-training an entire network can be computationally expensive for very deep networks. By correcting the most distortion-susceptible convolutional filter outputs, we are not only able to make a DNN robust to severe distortions, but are also able to maintain a very good performance on clean images.

Although we focus on image classification and object recognition, our proposed approach is not only generic enough to apply to a wide selection of tasks that use DNN models such as object detection \cite{rcnn},\cite{pascal} or semantic segmentation \cite{coco} but also other less commonly occurring noise types like adversarial noise.

\ifCLASSOPTIONcaptionsoff
  \newpage
\fi




\bibliographystyle{IEEEtran}
\vspace*{-2ex}
\bibliography{myrefs}
\end{document}